\documentclass{article}

\usepackage{microtype}
\usepackage{graphicx}
\usepackage{booktabs} %

\usepackage{hyperref}

\usepackage{url}            %
\usepackage{booktabs}       %
\usepackage{amsfonts}       %
\usepackage{nicefrac}       %

\usepackage{graphicx}
\usepackage{subcaption}
\usepackage{amsmath}
\usepackage{amsthm}
\usepackage{tikz}
\usepackage{filecontents}
\usepackage{algorithm}
\usepackage{algorithmic}
\usepackage{float}
\usepackage{lipsum}

\usepackage[accepted]{icml2020}

\newcommand{\beq}{\begin{equation}}
\newcommand{\eeq}{\end{equation}}
\newcommand{\be}{\begin{equation}}
\newcommand{\ee}{\end{equation}}
\newcommand{\beqa}{\begin{eqnarray}}
\newcommand{\eeqa}{\end{eqnarray}}
\newcommand{\bean}{\begin{eqnarray*}}
\newcommand{\eean}{\end{eqnarray*}}

\theoremstyle{definition}

\newcommand{\cO}{{\mathcal O}}

\newcommand{\cS}{{\mathcal S}}

\newcommand{\cX}{{\mathcal X}}

\newcommand{\vc}{{\mathbf c}}
\newcommand{\vd}{{\mathbf d}}
\newcommand{\ve}{{\mathbf e}}
\newcommand{\vh}{{\mathbf h}}

\newcommand{\vp}{{\mathbf p}}
\newcommand{\vr}{{\mathbf r}}
\newcommand{\vs}{{\mathbf s}}
\newcommand{\vu}{{\mathbf u}}
\newcommand{\vx}{{\mathbf x}}
\newcommand{\vO}{{\mathbf O}}

\newcommand{\bR}{{\mathbb R}}
\newcommand{\bZ}{{\mathbb Z}}

\newcommand{\fI}{{\mathfrak I}}
\newcommand{\fO}{{\mathfrak O}}
\newcommand{\fo}{{\mathfrak o}}

\icmltitlerunning{S2RM: Spatially Structured Recurrent Modules}

\begin{document}

\twocolumn[
\icmltitle{Spatially Structured Recurrent Modules}

\icmlsetsymbol{equal}{*}

\begin{icmlauthorlist}
\icmlauthor{Nasim Rahaman}{mpi,mila}
\icmlauthor{Anirudh Goyal}{mila}
\icmlauthor{Muhammad Waleed Gondal}{mpi}
\icmlauthor{Manuel Wuthrich}{mpi}
\icmlauthor{Stefan Bauer}{mpi}
\icmlauthor{Yash Sharma}{bethgelab}
\icmlauthor{Yoshua Bengio}{mila}
\icmlauthor{Bernhard Sch\"olkopf}{mpi}
\end{icmlauthorlist}

\icmlaffiliation{mpi}{Max Planck Institute for Intelligent Systems, T\"ubingen}
\icmlaffiliation{mila}{Mila, Qu\'ebec}
\icmlaffiliation{bethgelab}{Bethgelab, University of T\"ubingen}

\icmlcorrespondingauthor{Nasim Rahaman}{nasim.rahaman@tuebingen.mpg.de}

\icmlkeywords{Machine Learning, ICML}

\vskip 0.3in
]

\printAffiliationsAndNotice{\icmlEqualContribution} %

\begin{abstract}
Capturing the structure of a data-generating process by means of appropriate inductive biases can help in learning models that generalize well and are robust to changes in the input distribution. While methods that harness spatial and temporal structures find broad application, recent work \citep{goyal2019recurrent} has demonstrated the potential of models that leverage sparse and modular structure using an ensemble of sparingly interacting modules.
In this work, we take a step towards dynamic models that are capable of simultaneously exploiting both modular and spatiotemporal structures.
We accomplish this by abstracting the modeled dynamical system as a collection of autonomous but sparsely interacting \emph{sub-systems}. The sub-systems interact according to a topology that is learned, but also informed by the spatial structure of the underlying real-world system. 
This results in a class of models that are well suited for modeling the dynamics of systems that only offer \emph{local views} into their state, along with corresponding spatial locations of those views. 
On the tasks of video prediction from cropped frames and multi-agent world modeling from partial observations in the challenging Starcraft2 domain, we find our models to be more robust to the number of available views and better capable of generalization to novel tasks without additional training, even when compared against strong baselines that perform equally well or better on the training distribution. 
\end{abstract}

\section{Introduction} \label{sec:intro}

Many spatiotemporal complex systems can be abstracted as a collection of autonomous but sparsely interacting sub-systems, where sub-systems tend to interact if they are in each others' \emph{local vicinity} in some sense.
As an illustrative example, consider a grid of traffic intersections, wherein traffic flows from a given intersection to the adjacent ones, and the actions taken by some ``agent", say an autonomous vehicle, may at first only affect its immediate surroundings. Now suppose we want to forecast the future state of the traffic grid (say for the purpose of avoiding traffic jams). 

\begin{figure*}
\begin{subfigure}[t]{0.33\textwidth} 
\centering
\includegraphics[width=1\textwidth]{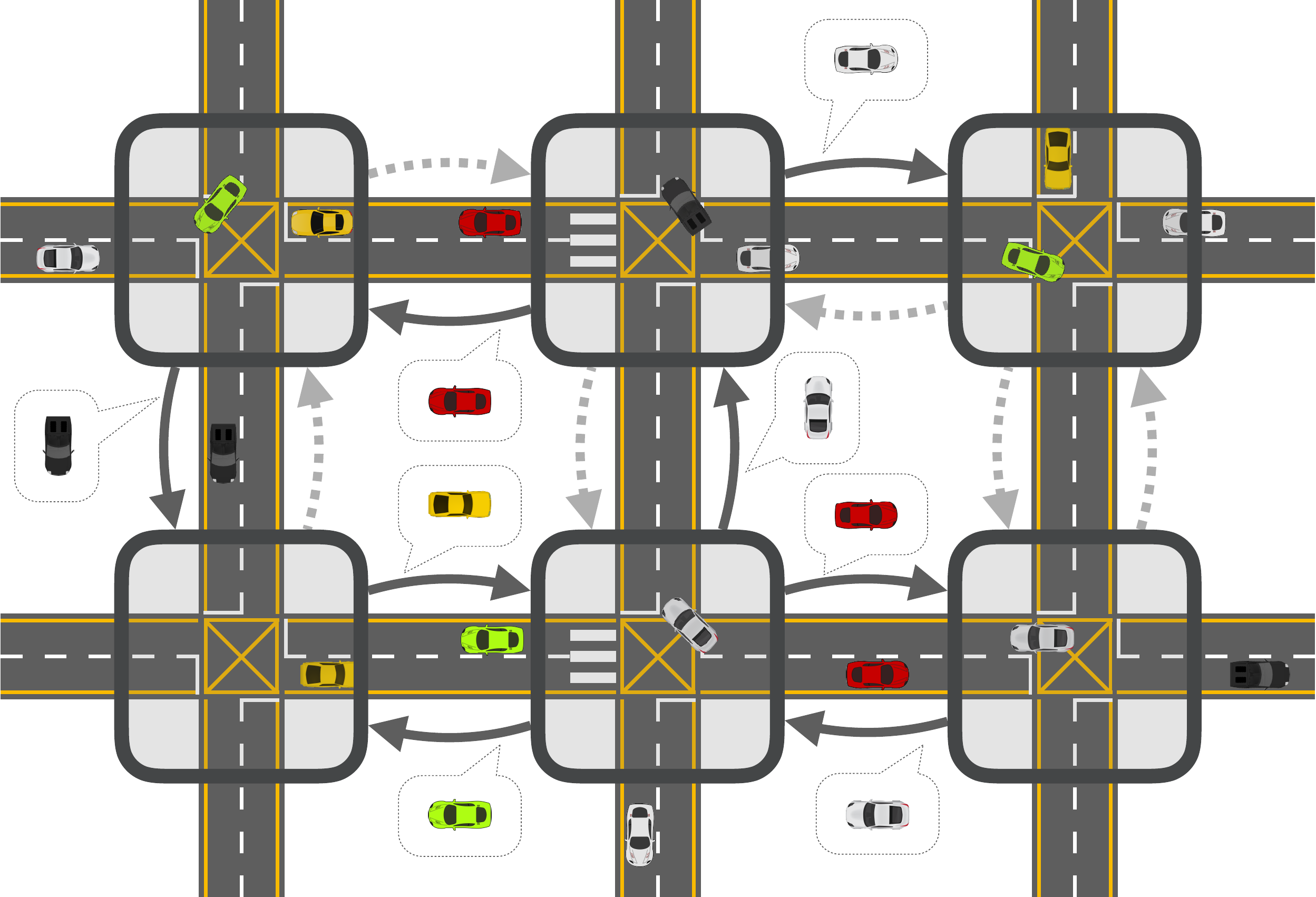}
\caption{\small Fully localized sub-systems.} \label{fig:one_interacting}
\end{subfigure}
\hfill
\begin{subfigure}[t]{0.33\textwidth}
\centering
\includegraphics[width=1\textwidth]{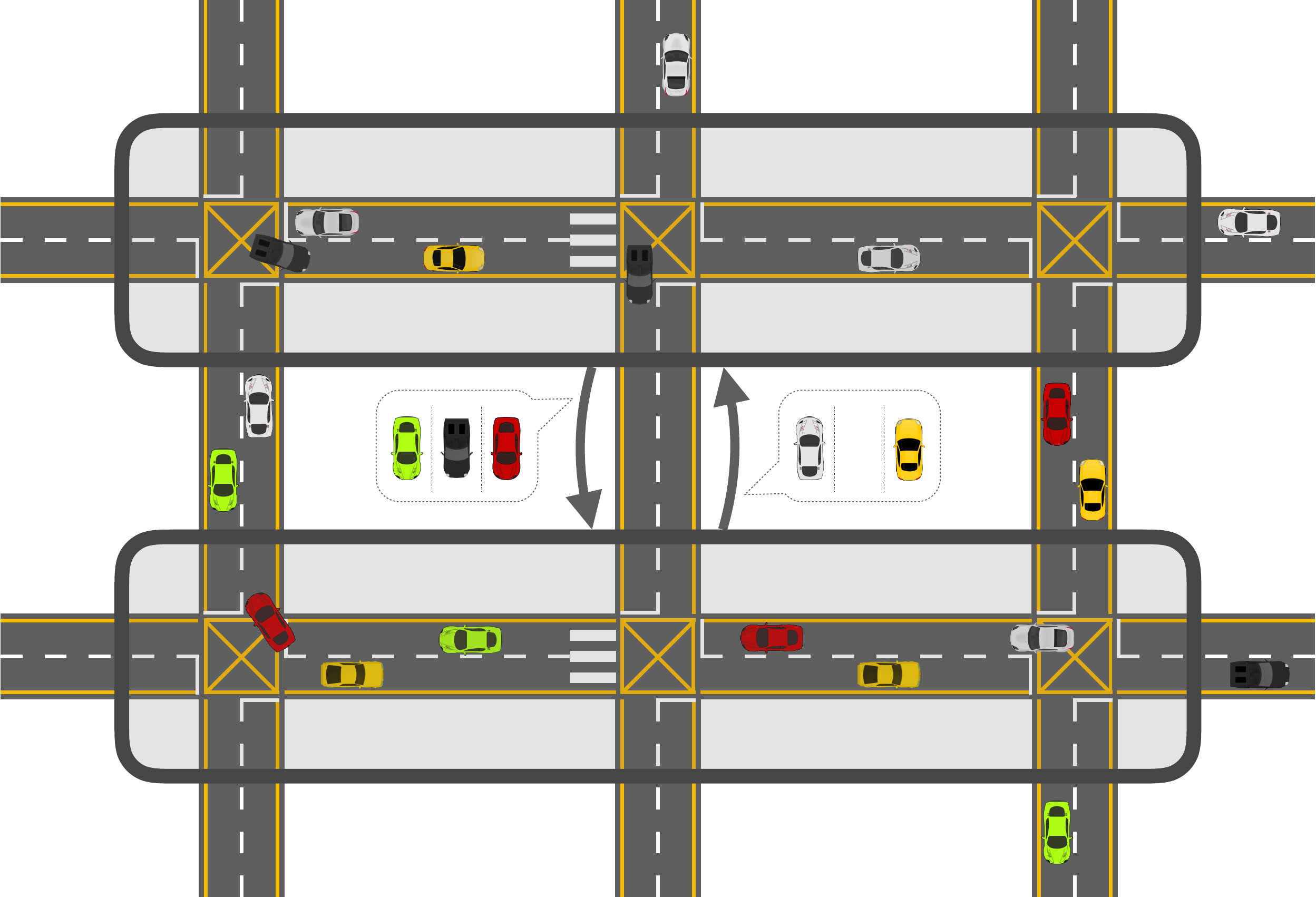}
\caption{\small Middle ground.} \label{fig:one_seminteracting}
\end{subfigure} 
\hfill
\begin{subfigure}[t]{0.33\textwidth}
\centering
\includegraphics[width=1\textwidth]{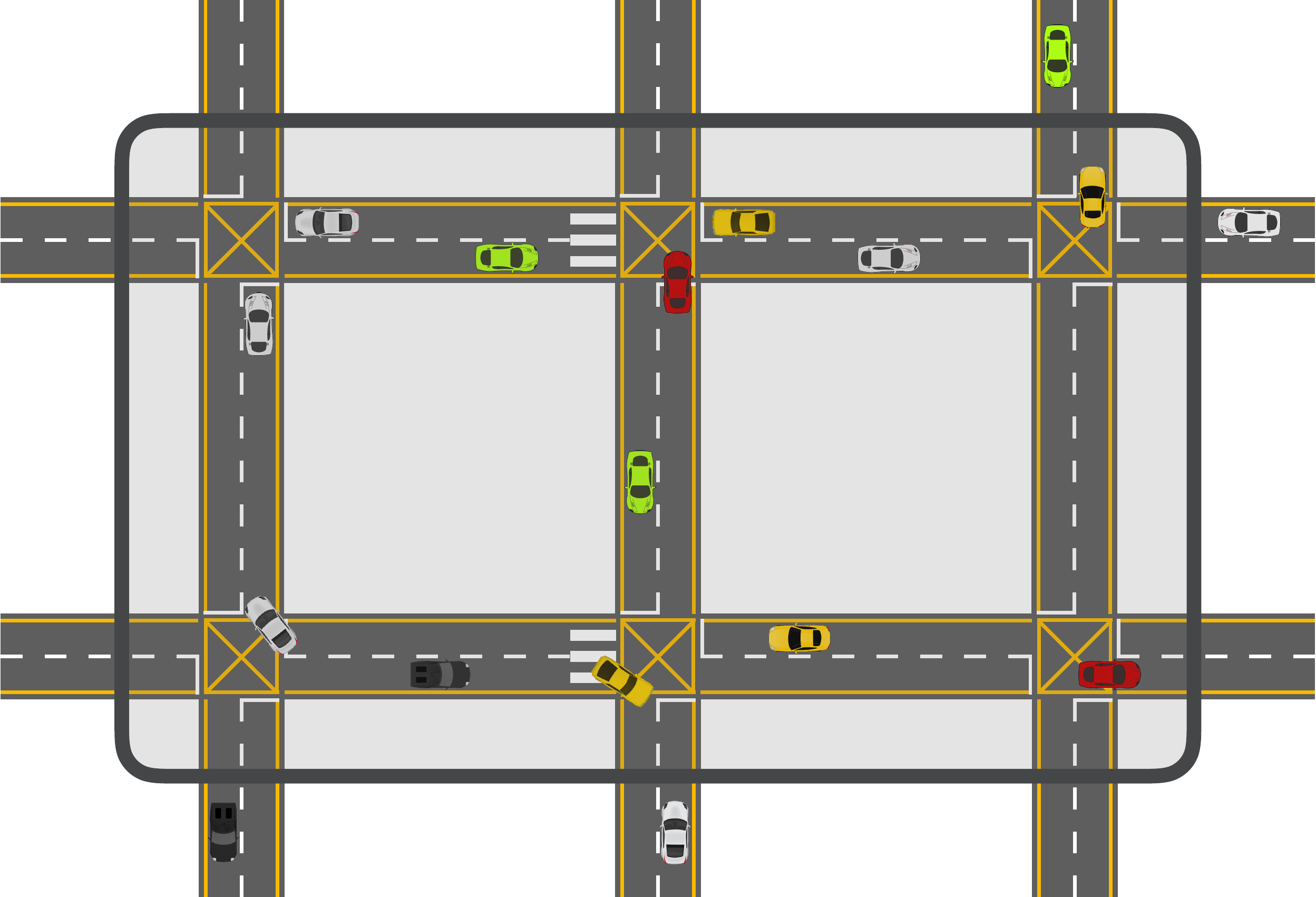}
\caption{\small Single, monolithic system.} \label{fig:one_noninteracting}
\end{subfigure}
\caption{\small A schematic representation of the spectrum of modeling strategies. Solid arrows with speech bubbles denote (dynamic) messages being passed between \emph{sub-systems} (dotted arrows denote the lack thereof). \textbf{Gist:} on the one end of the spectrum, (Figure~\ref{fig:one_interacting}), we have the strategy of abstracting each intersection as a sub-system that interact with neighboring sub-systems. On the other end of the spectrum (Figure~\ref{fig:one_noninteracting}) we have the strategy of modeling the entire grid with one monolithic system. The middle ground (Figure~\ref{fig:one_seminteracting}) we explore involves letting the model \emph{develop} a notion of locality by (say) abstracting entire avenues with a single sub-system.} \label{fig:one}
\vspace{-10pt}
\end{figure*}

There is a spectrum of possible strategies to model the system at hand. On one end of it lies the most general strategy: namely, one that calls for considering the entirety of all intersections simultaneously to predict the next state of the grid (Figure~\ref{fig:one_noninteracting}). The resulting model class can in principle account for interactions between any two intersections, irrespective of their spatial distance. However, the number of interactions such models must consider does not scale well with the size of the grid, and the strategy might be rendered infeasible for large grids with hundreds of intersections. 
 
On the other end of the spectrum is a specialized strategy that involves abstracting the dynamics of each intersection as an autonomous \emph{sub-system}, and having each sub-system interact only with other sub-systems associated with the four (or more) neighboring intersections (Figure~\ref{fig:one_noninteracting}). The interactions may manifest as messages that one sub-system passes to another and possibly contain information about how many vehicles are headed towards which direction, resulting in a collection of message passing entities (i.e., sub-systems) that collectively model the entire grid. By adopting this strategy, one assumes that the immediate future of any given intersection is affected only by the present states of the neighboring intersections, and not some intersection at the opposite end of the grid. The resulting class of models scales well with the size of the grid, but is possibly unable to model certain long-range interactions that could be leveraged to efficiently distribute traffic flow.

The spectrum above parameterizes the extent to which the spatial structure of the underlying system being modeled is incorporated into the design of the model. The former extreme ignores spatial structure altogether, resulting in a class of models that can be expressive but whose sample and
computational complexity do not scale well with the size of the system. The latter extreme results in a class of models that can scale well, but its adequacy (in terms of expressivity) is contingent on a predefined notion of locality (in the example above: the immediate four-neighborhood of an intersection). In this work, we aim to explore a middle-ground between the two extremes: namely, by proposing a class of models that does leverage the spatial structure, but by developing a notion of locality instead of relying on a predefined one (Figure~\ref{fig:one_seminteracting}). Reconsidering the traffic grid example: the proposed strategy results in a model that can potentially learn to abstract (say) entire avenues with a single sub-system. The interactions between intersections are therefore replaced by those between avenues, resulting in a scheme where a single sub-system might account for events that are spatially distant (such as those in the opposite ends of an avenue), but two events that are spatially closer together (such as those on two adjacent avenues of the same street, where streets run perpendicular to avenues) might be accounted for by different sub-systems. 

To implement this scheme, we will model the sub-systems as independent recurrent neural networks (RNNs) that interact sparsely via a bottleneck of attention \citep{goyal2019recurrent}, but extend this idea along two salient dimensions. First, we relax the assumption that the interaction topology between sub-systems (i.e., RNNs) is all-to-all, in the sense that all sub-systems are allowed to interact with all other sub-systems. We achieve this by learning to embed each sub-system in an embedding space endowed with a metric, and attenuate the interaction between two given sub-systems by their distance in this space (i.e., sub-systems too far away from each other in this space are not allowed to interact). Second, instead of assuming that the entire system is perceived simultaneously, we only assume access to \emph{local} (partial) observations alongside with the associated \emph{spatial locations}, resulting in a setting that partially resembles that of \citet{eslami2018neural}. Expressed in the language of the example above: we do not expect a birds eye view of the traffic grid, but only (say) LIDAR observations from autonomous vehicles at known GPS coordinates, or video streams from traffic cameras at known locations. The spatial location associated with an observation plays a crucial role in the proposed architecture in that we map it to the embedding space of sub-systems and \emph{address} the corresponding observation only to sub-systems whose embeddings lie in close vicinity. Likewise, to predict future observations at a queried spatial location, we again map said location to the embedding space and poll the states of sub-systems situated nearby. The result is a model that can \emph{learn} which spatial locations are to be associated with each other and be accounted for by the same sub-system. As an added plus, the parameterization we obtain is not only agnostic to the number of available observations and query locations, but also to the number of sub-systems. 

To evaluate the proposed model, we choose a problem setting where \textbf{(a)} the task is composed of different sub-systems or \emph{processes} that locally interact both spatially and temporally, and \textbf{(b)} the environment offers local views into its state paired with their corresponding spatial locations. The challenge here lies in building and maintaining a consistent representation of the \emph{global state} of the system given only a set of partial observations. To succeed, a model must learn to efficiently capture the available observations and place them in appropriate spatial context. The first problem we consider is that of video prediction from crops, analogous to that faced by visual systems of many animals: given a set of small crops of the video frames centered around stochastically sampled pixels (corresponding to where the fovea is focused), the task is to predict the content of a crop around \emph{any} queried pixel position at a future time. The second problem is that of multi-agent world modeling from partial observations in spatial domains, such as the challenging Starcraft2 domain \citep{samvelyan2019starcraft, vinyals2017starcraft}. The task here is to model the dynamics of the \emph{global} state of the environment given \emph{local} observations made by cooperating agents and their corresponding actions. Importantly and unlike prior work \citep{sun2018predicting}, our parameterization is agnostic to the number of agents in the environment, which can be flexibly adjusted on the fly as new agents become available or existing agents retire. This is beneficial for generalization in settings where the number of agents during training and testing are different.

\textbf{Contributions.} \textbf{(a)} We propose a new class of models, which we call Spatially Structured Recurrent Modules or S2RMs, which perform attention-driven spatially local modular computations. \textbf{(b)} We evaluate S2RMs (along with several strong baselines) on a selection of challenging problems to find that S2RMs are robust to the number of available observations and can generalize to novel tasks. 

\section{Problem Statement} \label{sec:problem_statement}
In this section, we build on the intuition from the previous section to formally specify the problem we aim to approach with the methods described in the later sections.

Let $\cX$ be a metric space, $\cO$ some set of possible \emph{observations}, and $\fO_{\cX}$ a set of mappings $\cX \to \cO$. Now, consider the \emph{evolution function} of a discrete-time dynamical system: 
\begin{align}
&\phi: \bZ \times \fO_{\cX} \to \fO_{\cX} \, \text{satisfying} \label{eq:dynamic_system_def}\\
&\phi(0, \fo) = \fo \; \text{where} \; \fo \in \fO_{\cX} \; \text{and} \nonumber \\
&\phi(t_2, \phi(t_1, \fo)) = \phi(t_1 + t_2, \fo) \; \text{for} \; t_1, t_2 \in \bZ \nonumber
\end{align} 
Informally, $\fo$ can be interpreted as the \emph{world state} of the system; together with a spatial \emph{location} $\vx \in \cX$, it gives the \emph{local} observation $\vO = \fo(\vx) \in \cO$. Given an initial world state $\fo$, the mapping $\phi(t, \fo)$ yields the world state at some (future) time $t$, thereby characterizing the dynamics of the system (which might be stochastic). 

While the above class of dynamical systems is fairly general, we now place a crucial restriction: namely, that for any pair of space-time \emph{events} $(t_1, \vx_1) \in (\bZ \times \cX)$ and $(t_2, \vx_2)$ and given any initial state $\fo_0 \in \fO_{\cX}$, there exists a finite $C>0$ such that the observation $\phi(t_1, \fo_0)(\vx_1)$ can influence the observation $\phi(t_2, \fo_0)(\vx_2)$ only if $t_2 \ge t_1$ and $d_{\cX}(\vx_1, \vx_2) \le C \cdot (t_2 - t_1)$, where $d_{\cX}$ is the metric on $\cX$. This assumption induces a notion of spatio-temporal locality by imposing that the effect of any given event can only propagate at a finite speed, where the latter is upper bounded by $C$. 

In this work, we are concerned with modelling systems that are subject to the above restriction. Assuming the system satisfies said restriction, we have the following

\textbf{Problem:} At every time step $t = 0, ..., T$, we are given a set of positions $\{\vx^{a}_{t}\}_{a = 1}^{A}$ and the corresponding observations $\{\vO_{t}^{a}\}_{a = 1}^{A}$, where  $\vO_{t}^{a} := \phi(t, \fo_0)(\vx^{a})$ for some initial world state $\fo_0$. The task is to infer the world state $\phi(t', \fo_0)$ at some future time-step $t' > T$.

In the traffic grid example of Section~\ref{sec:intro}, one could imagine $a$ as indexing traffic cameras or autonomous vehicles (i.e., \emph{observers}), $\vx_{t}^{a}$ as the GPS coordinates of observer $a$, and $\vO_{t}^{a}$ as the corresponding sensor feed (e.g. LIDAR observations or video streams from vehicles or traffic cameras). 

\section{Modelling Assumptions} \label{sec:assumptions} \label{sec:method}

Given the problem in Section~\ref{sec:problem_statement}, we now constrain it by placing certain structural assumptions. These assumptions will ultimately inform the inductive biases we select for the model (proposed in Section~\ref{sec:s2rm}); nevertheless, we remark beforehand that as with any inductive bias, their applicability is subject to the properties of the system being modeled and the objectives\footnote{E.g. generalization, sample complexity, robustness, etc.} being optimized. 

\textbf{Recurrent Dynamics Modeling.} While there exist multiple ways of modeling dynamical systems, we shall focus on recurrent neural networks (RNNs). Typically, RNN-based dynamics models are expressed as functions of the form: 
\beq
\vh_{t + 1} = F(\vO_t, \vh_{t}) \qquad \vO_t = D(\vh_t)
\eeq
where $\vO_t$ is the observation at time $t \in \bZ$, and $\vh_{t + 1}$ is the hidden state of the model. $F$ can be thought of as the parameterized \emph{forward-evolution function} the hidden state $\vh$ conditioned on the observation $\vO$, whereas $D$ is a \emph{decoder} that maps the hidden state to observations. Here, the evolution function of the modelled dynamical system (as defined in Equation~\ref{eq:dynamic_system_def}) can be obtained by rolling out the forward-evolution function in time. 

\textbf{Decomposition into Sub-systems.} Without loss of generality, one may assume that the dynamical system $\phi$ defined in Equation~\ref{eq:dynamic_system_def} can be decomposed into constituent systems $(\phi_1, \phi_2, ..., \phi_M)$, such that the interaction between all pairs of sub-systems $(\phi_i, \phi_j)$ satisfy some criterion. Now, the strength of this assumed criterion lies on a spectrum. On one end of the spectrum is the case where such a criterion is non-existent, i.e., no such decomposition is assumed and full generality is restored; this is the modeling assumption made when using conventional recurrent models like GRUs \citep{cho2014learning}, LSTMs \citep{hochreiter1997long} and vanilla RNNs. On the other end of the spectrum lies a setting where the decomposition is required to be such that sub-systems do not interact, i.e., $\phi_i$ and $\phi_j$ have independent dynamics \citep{li2018independently}. \citet{goyal2019recurrent} explore a middle ground, where the interaction between sub-systems $(\phi_i, \phi_j)$ are possible but constrained. In particular, they investigate a setting where the sub-systems are assumed to interact sparsely, and the interaction pattern (i.e., which sub-systems interact with which others) is dynamic and may depend on the world state $\fo$. In this work, we adopt the assumption of sparsely interacting sub-systems, but subject the interaction pattern to an additional spatial constraint. 

\textbf{Local Interactions Between Sub-systems.} In addition to assuming dynamic sparse interactions between sub-systems, we also assume that a given sub-system $\phi_j$ may \emph{preferentially} interact with another given sub-system $\phi_i$. Intuitively, one may think of $\phi_j$ as lying in \emph{vicinity} of $\phi_i$. This naturally leads us to a notion of topology over sub-systems, one where sub-systems situated in each other's local neighborhood are less constrained in their interactions. In the next section, we will discuss how we model this topology by associating each sub-system $\phi_i$ with a learned embedding $\vp_i$ in an existing metric space, which we will call $\cS$. Subsequently, the affinity of sub-system $\phi_i$ to interact with another sub-system $\phi_j$ will be quantified by a \emph{similarity measure} $Z$, such that $Z(\vp_i, \vp_j)$ is large if $\phi_i$ and $\phi_j$ prefer to interact. 

\textbf{Locality of Observations.} Recall from Section~\ref{sec:problem_statement} that the observations available to the model respect a notion of spatio-temporal locality. However, this notion of locality is distinct from the one between sub-systems (induced via $Z$), and one important modeling decision is how the two should interact. We propose to embed the position $\vx \in \cX$ associated with an observation $\vO$ to the metric space of sub-systems $\cS$ via a continuous and one-to-one mapping $P: \cX \to \cS$, which allows us to \emph{match} the observation $\vO$ to all sub-systems $\phi_m$ in the vicinity of $P(\vx) \in \cS$, i.e., where $Z(P(\vx), \vp_m)$ is sufficiently large. Likewise, the same sub-systems $\phi_m$ are polled if the model is queried for a prediction at $\vx$. On a high level, this results in a scheme where each subsystem $\phi_m$ can account for observations made at a set of positions $\cX_m \subset \cX$, which we call its \emph{enclave}. In particular, the enclaves $\cX_i$ and $\cX_j$ corresponding to sub-systems $\phi_i$ and $\phi_j$ may overlap, and we do not constrain the distance between two given points in $\cX_m$ to be small.

\section{Proposed Model} \label{sec:s2rm}

\begin{figure} 
\centering
\includegraphics[width=0.49\textwidth]{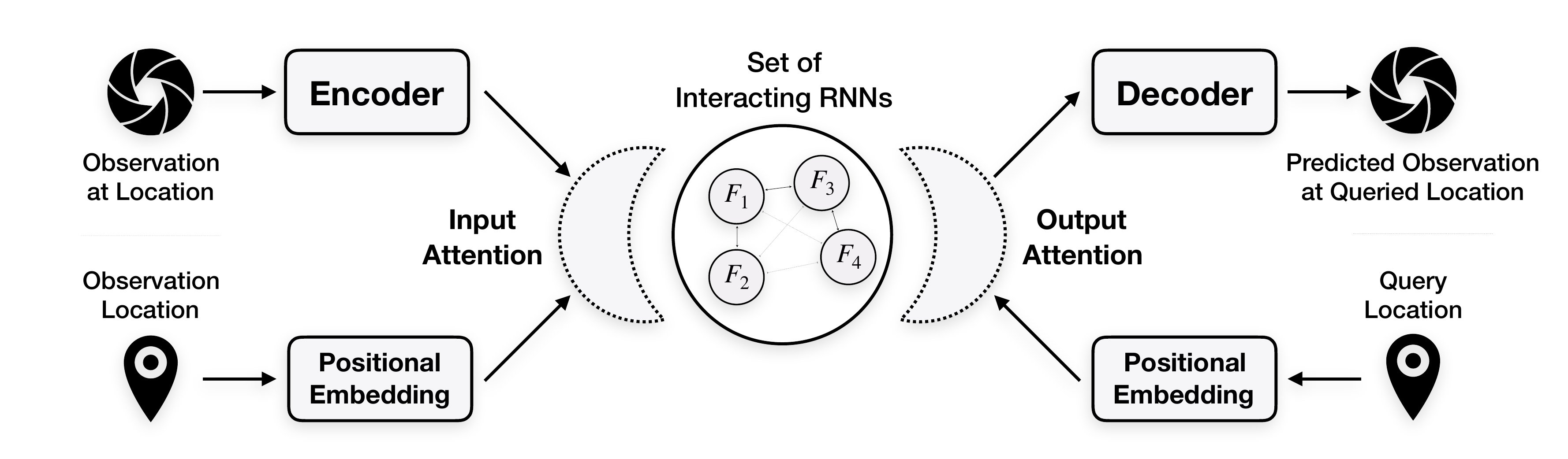}
\caption{Schematic representation of the proposed architecture.}
\label{fig:arch}
\vspace{-15pt}
\end{figure}

Informed\footnote{In doing so, we use the assumptions merely as guiding principles; we do not claim that we \emph{infer} e.g. the true decomposition of the ground-truth system, even if all assumptions are satisfied.} by the model assumptions detailed in the previous section, we now proceed to describe the proposed model -- Spatially Structured Recurrent Modules or \textbf{S2RM} -- which comprise the following components (Figure~\ref{fig:arch}):

\textbf{Model Inputs.} Recall from Section~\ref{sec:problem_statement} that we have for every time step $t = 0, ..., T$ a set of tuples of positions and observations $\{(\vx_{t}^{a}, \vO_{t}^{a})\}_{a = 1}^{A}$ where $\vx_{t}^{a} \in \cX$ and $\vO_{t}^{a} \in \cO$ for all $t$ and $a$. To simplify, we assume that $\cX \subset \bR^n$, and denote by $x_i$ the $i$-th component of the vector $\vx \in \cX$. 
\textbf{Encoder.} The encoder $E$ is a parameterized function mapping observations $\vO$ to a corresponding vector representation $\ve = E(\vO)$. Here, $E$ processes all observations in parallel across $t$ and $a$ to yield representations $\ve_{t}^{a}$. 

\textbf{Positional Embedding.} The positional embedding $P$ is a fixed mapping from $\cX$ to $\cS$. We choose $\cS$ to be the unit sphere in $d$-dimensions, $d$ being a multiple of $2n$, and the positional encoder as the following function:
\begin{align}
P(\vx) &= \nicefrac{\vs}{\|\vs\|} \in \cS \quad \text{where} \\
s_{2i + m} = \sin{\left(\nicefrac{x_m}{10000^i}\right)} \; & \; s_{2i + 1 + m} = \cos{\left(\nicefrac{x_m}{10000^i}\right)} 
\end{align}
with $m = 1, ..., n$ and $i = 1, ..., \nicefrac{d}{2n}$. While the above function is commonly used \citep{vaswani2017attention}, other choices might also be viable. Accommodating a slight abuse of notation, we will refer to $P(\vx)$ as $\vs$ and $P(\vx_{t}^{a})$ as $\vs_t^a$.

\textbf{Set of Interacting RNNs.} To model the dynamics of the world state, we use a set of $M$ independent RNN \emph{modules} (like in ~\citet{goyal2019recurrent}), which we denote as $\{F_m\}_{m = 1}^{M}$. To each $F_m$, we associate an embedding vector $\vp^m \in \cS$, where all $\{\vp^m\}_{m = 1}^{M}$ are learnable parameters. On a high level, RNNs $F_m$ interact with each other via an \emph{inter-cell attention}, and with the input representations $\ve_{t}^{a}$ via \emph{input attention}. More precisely, at a given time step $t$, each $F_m$ expects an input $\vu^m_t$, together with an \emph{aggregated hidden state} $\bar\vh^m_t$ and optionally a memory state $\vc^m_t$ to yield the hidden and memory states at the next time step: 
\beq \label{eq:rnns}
(\vh_{t+1}^m, \vc_{t+1}^m) = F_m(\vu_t^m, \bar\vh_t^m, \vc_t^m)
\eeq
where the input $\vu_t^m$ results from the input attention and $\bar\vh_t^m$ from the inter-cell attention (both described below). If available, the memory state $\vc_t^m$ resembles the cell state in an LSTM \citep{hochreiter1997long}.

\textbf{Input Attention.} Similar to MHDPA (multi-head dot-product attention, \citet{vaswani2017attention}), the input attention mechanism is a mapping between sets: namely, from that of observation encodings $\{\ve_t^{a}\}_{a=1}^{A}$ to that of RNN inputs $\{\vu_t^m\}_{m=1}^{M}$. In what follows, we use the einsum notation\footnote{Indices not appearing on both sides of an equation are summed over; this is implemented as \texttt{einsum} in most DL frameworks.} to succintly describe the exact mechanism. But before that, we define the truncated spherical Gaussian kernel \citep{fasshauer2011positive} to quantify the similarity between two points $\vp, \vs \in \cS$: 
\begin{align} \label{eq:zonalkernel}
Z(\vp, \vs) = \begin{cases}
      \exp{\left[-2\epsilon (1 - \vp \cdot \vs)\right]}, & \text{if}\ \vp \cdot \vs \ge \tau \\
      0, & \text{otherwise}
\end{cases}
\end{align}
where $\epsilon \in \bR^{+}$ and $\tau \in [-1, 1)$ are hyper-parameters (\emph{kernel bandwidth} and \emph{truncation parameter}, respectively), and $0 \le Z \le 1$ since $\vp$ and $\vs$ are unit vectors. Now, we use $k$ to index the attention heads, $d$ to index the dimension of the key and query vectors, and denote with $e_{ai}$ the $i$-th component of $\ve^a_t$ and with $h_{mj}$ the $j$-th component of $\vh^m_t$. Given learnable parameters $\Theta^{(K)}$, $\Theta^{(Q)}$, $\Theta^{(V)}$, we obtain:
\begin{align} \label{eq:input_attn_0}
Q_{akd} = e_{ai} \Theta_{ikd}^{(Q)} \quad&\quad K_{mkd} = h_{mj} \Theta_{jkd}^{(K)} \\
V_{akv} = e_{ai} \Theta_{ikv}^{(V)} \quad&\quad \tilde{W}_{mak} = Q_{akd} K_{mkd} \\
\bar{W}_{mak} = \text{sm}_{a} (\tilde{W}_{mak}) \quad&\quad W^{(L)}_{ma} = Z(\vp^m, \vs^a) \\
W_{mak} = W^{(L)}_{ma} \bar{W}_{mak} \quad&\quad \tilde{u}_{m(kv)} = W_{mak} V_{akv} \label{eq:input_attn_2}
\end{align}
where: $\text{sm}_{a}$ denotes softmax along the $a$-dimension, $W^{(L)}$ is what we will call the \emph{local weights}, we omit the time subscript in $\vs^a$ for notational clarity, and $\tilde{u}_{m(kv)}$ is the $(kv)$-th component of a vector $\tilde\vu^m$. Finally, we obtain the components $u_{mi}$ of RNN inputs $\vu^m_t$ via a gating operation: 
\begin{align}
u_{mi} = G^{\text{(inp)}}_m \cdot b_{mi} + (1 - G^{\text{(inp)}}_m) \cdot \tilde{u}_{mi}
\end{align}
where the gating weight $G^{\text{(inp)}}_m \in (0, 1)$ is obtained by passing $\tilde u_{mi}$ and $b_{mi} = W^{(L)}_{ma}e_{ai}$ through a two-layer MLP with sigmoidal output (in parallel across $m$). Now, observe that by weighting the MHDPA attention outputs ($\bar{W}$ in Equation~\ref{eq:input_attn_2}) by the kernel $Z$ (via $W^{(L)}$), we construct a scheme where the interaction between input $\vO_{t}^{a}$ and RNN $F_m$ is allowed only if the embedding $\vs^a_t$ of the corresponding position $\vx_t^a$ has a large enough cosine similarity ($\ge \tau$) to the embedding $\vp_m$ of $F_m$. This partially implements the assumption of \emph{Locality of Observation} detailed in Section~\ref{sec:assumptions}.

\textbf{Inter-cell Attention.} The inter-cell attention maps the hidden states of each RNN $\{\vh_t^m\}_{m=1}^{M}$ to the set of \emph{aggregated hidden states} $\{\bar\vh_t^m\}_{m=1}^{M}$, thereby enabling interaction between the RNNs $F_m$. While its mechanism is identical to that of the input attention, we formulate it below for completeness. To proceed, we denote with $h_{li}$ the $i$-th component of $\vh_t^l$ (in addition to the notation introduced before Equation~\ref{eq:input_attn_0}), and take $\Phi^{(Q)}$, $\Phi^{(K)}$ and $\Phi^{(V)}$ to be learnable parameters. We have:
\begin{align} \label{eq:intercell_attn_0}
Q_{mkd} = h_{mj} \Phi_{jkd}^{(Q)} \quad&\quad K_{lkd} = h_{li} \Phi_{ikd}^{(K)} \\
V_{lkv} = h_{li} \Phi_{ikv}^{(V)} \quad&\quad \tilde{W}_{mlk} = Q_{mkd} K_{lkd} \\
\bar{W}_{mlk} = \text{sm}_{l} (\tilde{W}_{mlk}) \quad&\quad W^{(L)}_{ml} = Z(\vp^m, \vp^l) \\
W_{mlk} = \bar{W}_{mlk} W^{(L)}_{ml} \quad&\quad \tilde{h}_{m(kv)} = W_{mlk} V_{lkv} \label{eq:intercell_attn_3}
\end{align}
where $\tilde{h}_{m(kv)}$ is the $(kv)$-th component of a vector $\tilde\vh^m$. Finally, the $j$-th component $\bar h_{mj}$ of the aggregated hidden state $\bar \vh_t^m$ in Equation~\ref{eq:rnns} is given by a gating operation: 
\begin{align}
\bar h_{mj} = G^{\text{(ic)}}_m \cdot c_{mj} + (1 - G^{\text{(ic)}}_m) \cdot \tilde h_{mj}
\end{align}
where the gating weight $G^{\text{(ic)}}_m \in (0, 1)$ is obtained by passing $\tilde h_{mj}$ and $c_{mj} = W^{(L)}_{ml}h_{lj}$ through a two-layer MLP with sigmoid output (in parallel across $m$).
The weighting by $Z$ (in Equation~\ref{eq:intercell_attn_3}, left) ensures that the interaction is constrained to be only between RNNs whose embeddings in $\cS$ are similar enough, thereby implementing the assumption of \emph{Local Interactions between Sub-systems} in Section~\ref{sec:assumptions}. 

\textbf{Output Attention.} The output attention mechanism together with the decoder (described below) serve as an apparatus to evaluate the world state modeled (implicitly) by the set of RNNs ($\{F_m\}_{m=1}^{M}$) at time $t + 1$ (for one-step forward models). Given a query location $\vx^q \in \cX$ and its corresponding embedding $\vs^q = P(\vx^q) \in \cS$, the output attention mechanism polls the RNNs $F_m$ whose embeddings $\vp^m$ are similar enough to $\vs^q$, as measured by the kernel $Z$. Denoting $h_{mj}$ the $j$-th component of $\vh^{m}_{t + 1}$, we have: 
\beq
d^q_{j} = Z(\vs^q, \vp^m) \, h_{mj}
\eeq
where $d^q_{j}$ can be interpreted as the $j$-th component of the vector $\vd^{q}_{t + 1}$ associated with the query location $\vx^q$. 
    
\textbf{Decoder.} The decoder $D$ is a parameterized function that predicts the observation $\hat\vO_{t + 1}^q \in \cO$ at $\vx^q$ given the representation $\vd^{q}_{t + 1}$ from the output attention. 

This concludes the description of the generic architecture, which allows for flexibility in the choice of the RNN architecture (i.e., the internal architecture of $F_m$). In practice, we find Gated Recurrent Units (GRUs) \citep{cho2014learning} to work well, and call the resulting model Spatially Structured GRU or \textbf{S2GRU}. Moreover, Relational Memory Cores (RMCs) \citep{santoro2018relational} also profit from our architecture (with a minor modification detailed in Appendix~\ref{app:s2rmc}), and we refer to the resulting model as \textbf{S2RMC}.

\section{Related Work}

\textbf{Problem Setting.} Recall that the problem setting we consider is one where the environment offers local (partial) views into its global state paired with the corresponding spatial locations. With Generative Query Networks (GQNs), \citet{eslami2018neural} investigate a similar setting where the 2D images of 3D scenes are paired with the corresponding \emph{viewpoint} (camera position, yaw, pitch and roll). Given that GQNs are feedforward models, they do not consider the dynamics of the underyling scene and as such cannot be expected to be consistent over time \citep{kumar2018consistent}. \citet{singh2019sequential} and \citet{kumar2018consistent} propose variants that are temporally consistent, but unlike us, they do not focus on the problem of modeling the forward dynamics. 

\textbf{Modularity.} Modularity has been a recurring topic in the context of meta-learning \citep{alet2018modular, bengio2019meta, ke2019learning}, sequence modeling \citep{henaff2016tracking, goyal2019recurrent, li2018independently} and beyond \citep{jacobs1991adaptive, shazeer2017outrageously, parascandolo2017learning}. However, unlike prior work, we integrate modularity and spatio-temporal structure in a unified framework.  

\textbf{Spatial Attention.} Mechanisms for spatial attention have been well studied \citep{jaderberg2015spatial, wang2017nonlocal, zhang2018self, parmar2018image}, but they typically operate on image pixels. Our setting is more general in the sense that we do not necessarily require that the world state of the underlying system be represented by images. 

\textbf{Attention Mechanisms and Information Flow.} Attention mechanisms have been used to attenuate the flow of information between components of the network, e.g. in NTMs \citep{graves2014neural}, DNCs \citep{graves2016hybrid}, RMCs \citep{santoro2018relational}, SAB \citep{ke2018sparse} and Graph Attention Networks \citep{velivckovic2017graph, battaglia2018relational}. Our work contributes to this body of literature.

\section{Experiments} \label{sec:experiments}
In this section, we present a selection of experiments to empirically evaluate S2RMs and gauge their performance against strong baselines on two data domains. We proceed as follows: in Section~\ref{sec:baselines} we introduce the baselines, followed by experimental results on a video prediction task (Section~\ref{sec:bballs}) and on the multi-agent world modeling task in the challenging Starcraft2 domain (Section~\ref{sec:sc2}). Additional results and supporting plots can be found in Appendix~\ref{app:results}. 

\subsection{Baseline Methods} \label{sec:baselines}

\begin{figure*} 
\centering
\includegraphics[width=1\textwidth]{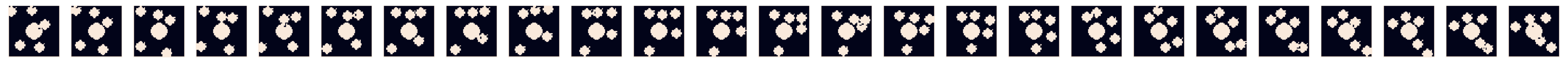}
\includegraphics[width=1\textwidth]{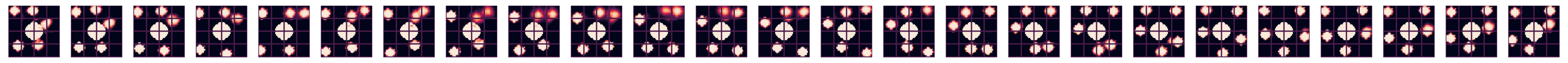}
\includegraphics[width=1\textwidth]{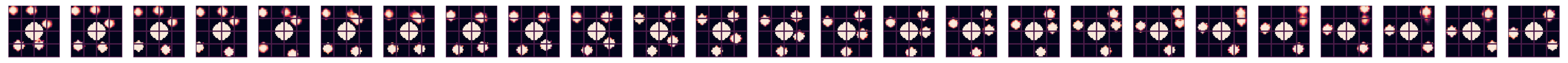}
\includegraphics[width=1\textwidth]{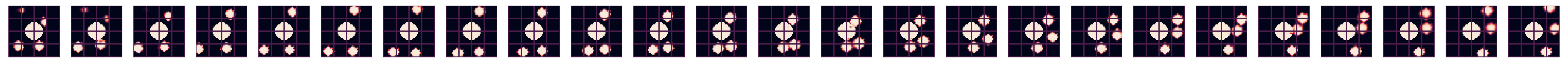}
\includegraphics[width=1\textwidth]{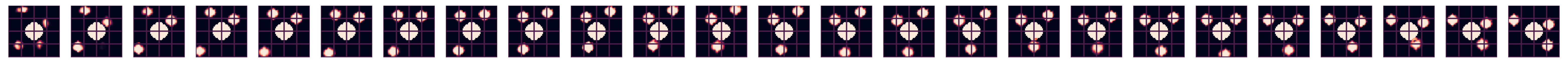}
\caption{Rollouts (OOD) with 5 bouncing balls, from top to bottom: ground-truth, S2GRU, RIMs, RMC, LSTM. Note that all models were trained on sequences with 3 bouncing balls, and the global state was reconstructed by stitching together $11 \times 11$ patches from the models (queried on a $4 \times 4$ grid). \textbf{Gist:} S2GRU succeeds at keeping track of all bouncing balls over long rollout horizons (25 frames).}
\label{fig:bb_roll}
\vspace{-10pt}
\end{figure*}

To draw fair comparisons, we require a baseline architecture that is agnostic to the number of observations $A$, is invariant to the ordering of $\{(\vx_t^a, \vO_t^a)\}_{a}^{A}$ with respect to $a$ and features a querying mechanism to extract a predicted observation $\vO^q_{t'}$ at a given query location $\vx^q$ in a future time-step $t' > t$. Fortunately, it is possible to obtain a performant class of models fulfilling our requirements by extending prior work on Generative Query Networks or GQNs \citep{eslami2018neural}. The resulting model has three components: 

\textbf{Encoder.} At a given timestep $t$, the encoder $E$ jointly maps the embedding $\vs_t^a \in \cS$ of the position $\vx_t^a \in \cX$ and the corresponding observations $\vO_t^a$ to encodings $\ve_t^a$, which are then summed over $a$ to obtain an \emph{aggregated representation}: 
\beq \label{eq:additive_agg}
\vr_t = \sum_{a = 1}^{A} E(\vO_t^a, \vs_t^a)
\eeq
The additive aggregation scheme we use is well known from prior work \citep{santoro2017simple, eslami2018neural, garnelo2018neural} and makes the model agnostic to $A$ and to permutations of $(\vx_t^a, \vO_t^a)$ over $a$. The encoder $E$ is a seven-layer CNN with residual layers, and the positional embedding $\vs_t^a$ is injected after the second convolutional layer via concatenation with the feature tensor. The exact architectures can be found in Appendices~\ref{app:encdec_arch_bb} and \ref{app:encdec_arch_sc2}. 

\textbf{RNN.} The aggregated representation $\vr_t$ is used as an input to a RNN model $F$ as following:  
\beq
\vh_{t+1}, \vc_{t+1} = F(\vr_t, \vh_t, \vc_t)
\eeq
where $\vh_t$ and $\vc_t$ are hidden and memory states of the RNN $F$ respectively. We experiment with various RNN models, including LSTMs \citep{hochreiter1997long}, RMCs \citep{santoro2018relational} and Recurrent Independent Mechanisms (RIMs) \citep{goyal2019recurrent}. 

As a sanity check, we also show results with a \emph{Time Travelling Oracle} (TTO), which has access to $r_{t + 1}$ (but at time step $t$), and produces $\vh_{t + 1} = F_{TTO}(\vr_{t+1})$ with a two layer MLP $F_{TTO}$. TTO therefore does not model the dynamics, 
but merely verifies that the additive aggregation scheme (Equation~\ref{eq:additive_agg}) and the querying mechanism (Equation~\ref{eq:context_query}) are sufficient for the task at hand. 

\textbf{Decoder.} Given the embedding $\vs^q$ of the query position $\vx^q$, the decoder $D$ predicts the corresponding observation $\hat\vO_{t+1}^q$:
\beq \label{eq:context_query}
\hat \vO_{t + 1}^q = D(\vh_{t + 1}, \vs^q)
\eeq
We parameterize $D$ with a deconvolutional network with residual layers, and inject the positional embedding of the query $\vs^q$ after a single convolutional layer by concatenating with the layer features (see Appendices~\ref{app:encdec_arch_bb} and \ref{app:encdec_arch_sc2}). 

The architecture described above therefore extends the framework of GQNs by predicting the forward dynamics of the aggregated representation; nevertheless, we do not consider it a novel contribution of this work. 

\subsection{Video Prediction from Crops} \label{sec:bballs}

\begin{figure} 
\centering
\includegraphics[width=0.48\textwidth]{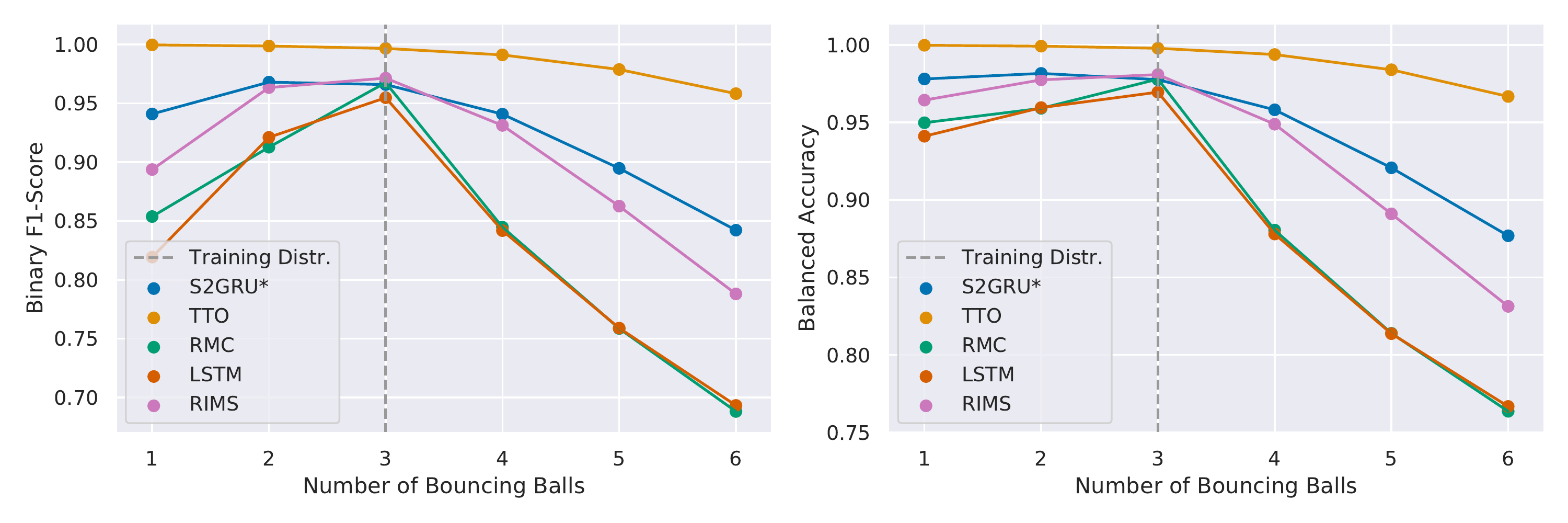}
\caption{Performance metrics on OOD one-step forward prediction task. \textbf{Gist:} S2GRU outperforms all RNN baselines OOD.}
\label{fig:bb_ood}
\vspace{-10pt}
\end{figure}

\textbf{Task Description.} We consider the problem of predicting the future frames of simulated videos of balls bouncing in a closed box \citep{miladinovic2019disentangled}, given only crops from the past video frames which are centered at known pixel positions. Using the notation introduced in Section~\ref{sec:problem_statement}: at every time step $t$, we sample $A = 10$ pixel positions $\{\vx_t^a\}_{a = 1}^{10}$ from the $t$-th full video frame $\fo_t$ of size $48 \times 48$. Around the sampled central pixel positions $\vx_t^a$, we extract $11 \times 11$ crops, which we use as the \emph{local} observations $\vO_t^a$. The task now is to predict $11 \times 11$ crops $\vO^{q}_{t'}$ corresponding to query central pixel positions $\vx^q_{t'}$ at a future time-step $t' > t$. Observe that at any given time-step $t$, the model has access to at most 52\% of the global video frame assuming that the crops never overlap (which is rather unlikely).

\begin{figure} 
\centering
\includegraphics[width=0.48\textwidth]{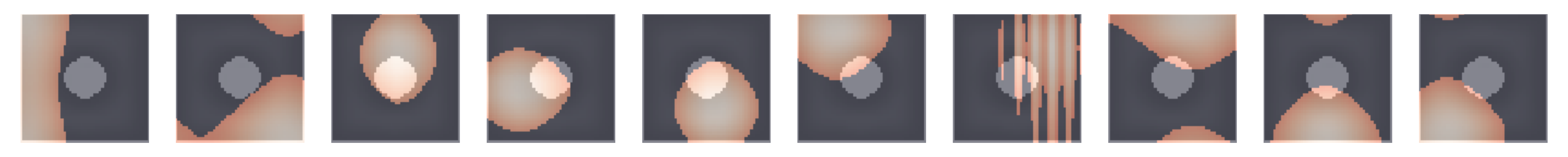}
\caption{Visualization of the spatial locations each module is responsible for modeling (i.e. the \emph{enclaves} $\cX_m$, defined in Section~\ref{sec:assumptions}). The central ball does not bounce, i.e. it is stationary in all sequences. \textbf{Gist:} the modules \emph{focus attention} on challenging regions, e.g. the corners of the arena and the surface of the fixed ball.}
\label{fig:bb_enclaves}
\vspace{-10pt}
\end{figure}

\textbf{Dataset.} We train all models on a training dataset of $20$K video sequences with $100$ frames of $3$ balls bouncing in an arena of size $48 \times 48$. We also include an additional fixed ball in the center to make the task more challenging. We use another $1$K video sequences of the same length and the same number of balls as a held-out validation set. In addition, we also have $5$ out-of-distribution (OOD) test sets with various number of bouncing balls (ranging from $1$ to $6$) and each containing $1$K sequences of length $100$. 

\textbf{Training.} We train all models until the validation loss is saturated, and select the best of three runs (more details in Appendix~\ref{app:hparams_train}). During training, we automatically decay the learning rate by a factor of $2$ if the validation loss does not decrease by at least $0.01\%$ for five consecutive epochs.

\textbf{Evaluation Criteria.} After having trained on the training dataset with 3 bouncing balls, we evaluate the performance on all test datasets with $1$ to $6$ bouncing balls. In Figure~\ref{fig:bb_ood}, we report the balanced accuracy (i.e. arithmetic mean of recall and specificity) and F1-scores (i.e. harmonic mean of precision and recall) to account for class-imbalance. Additionally, in Figure~\ref{fig:bb_roll}, we qualitatively show reconstructions from 25 step rollouts on the OOD dataset with 5 balls (see Appendix~\ref{app:results_bb}). Finally in Figure~\ref{fig:bb_enclaves}, we show for each module its corresponding enclave, which is the spatial region that it is \emph{responsible} for modelling, i.e. for pixels at position $\vx$, we plot $\{Z(P(\vx), \vp^m)\}_{m=1}^{10}$ (cf. Section~\ref{sec:s2rm}). 

\textbf{Results.} In Figure~\ref{fig:bb_ood}, we see that S2GRUs out-perform all non-oracle baselines OOD on the one-step forward prediction task and strike a good balance in regard to in-distribution and OOD performance. Note, however, that the additive aggregation scheme and querying mechanism (Equations~\ref{eq:additive_agg} and \ref{eq:context_query}) can indeed generalize, as shown by the good performance of the oracle (TTO). Figure~\ref{fig:bb_enclaves} shows how the modules \emph{share responsibility} of modelling the entire spatial domain, whereas Figure~\ref{fig:bb_roll} shows that S2GRUs can perform OOD rollouts over long horizons (25 frames) without losing track of balls. Additional results in Appendix~\ref{app:results_bb}. 

\subsection{Multi-Agent World Modeling on Starcraft2} \label{sec:sc2}
\begin{figure*} 
\centering
\includegraphics[width=1\textwidth]{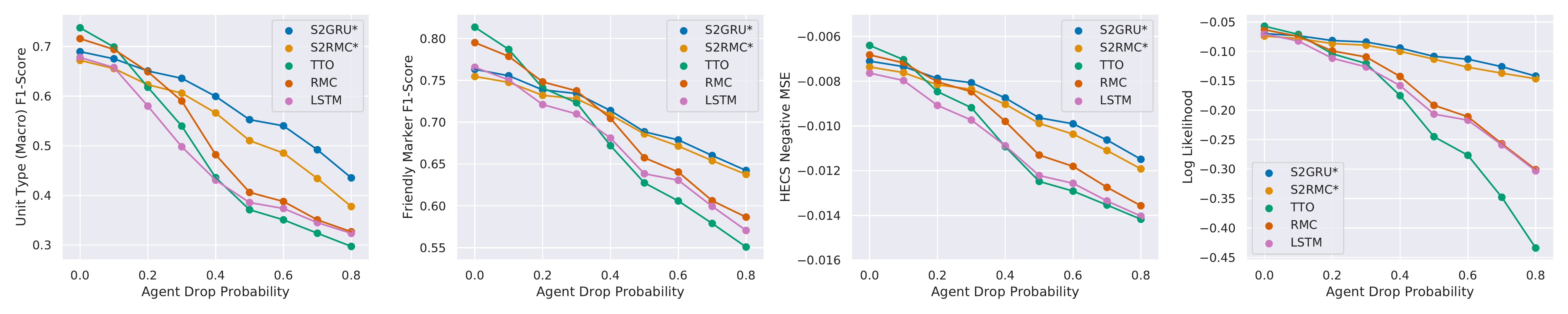}
\caption{Performance metrics (larger the better) as a function of the probability that an agent will not supply information to the world model but still query it. \textbf{Gist:} while all models lose performance as fewer agents share observations, we find S2RMs to be most robust.}
\label{fig:sc2_robustness}
\vspace{-10pt}
\end{figure*}
\begin{table}[]
    \centering
    \begin{tabular}{lrrrr}
    \toprule
    {}       &          UT-F1 &           FM-F1 &               NMSE &              LL \\
    \midrule
    \texttt{(1s2z)} &         &                 &                    &                 \\
    LSTM     &         0.6267 &          0.8464 &            -0.0040 &         -0.0382 \\
    RMC      &         0.6839 &          0.8597 &            -0.0033 &         -0.0334 \\
    S2GRU    &\textbf{0.7488} & \textbf{0.8627} &   \textbf{-0.0023} & \textbf{-0.0233}\\
    S2RMC    &         0.7317 &          0.8563 &            -0.0026 &         -0.0261 \\
    TTO      &         0.7518 &          0.8883 &            -0.0025 &         -0.0259 \\
    \midrule
    \texttt{(5s3z)} &         &                 &                    &                 \\
    LSTM     &          0.4975 &          0.7123 &          -0.0134 &          -0.1251 \\
    RMC      & \textbf{0.5414} & \textbf{0.7486} &          -0.0132 &          -0.1167 \\
    S2GRU    &          0.5310 &          0.7058 & \textbf{-0.0119} & \textbf{-0.1108} \\
    S2RMC    &          0.5114 &          0.6945 &          -0.0124 &          -0.1205 \\
    TTO      &          0.6115 &          0.7872 &          -0.0107 &          -0.0940 \\
    \bottomrule
    \end{tabular}
    \caption{Performance metrics on OOD scenarios \texttt{1s2z} and \texttt{5s3z} (larger numbers are better). The metrics are: unit-type macro F1 score (UT-F1), friendly-marker F1 score (FM-F1), HECS Negative Mean Squared Error (NMSE) and Log Likelihood (LL).}
    \label{tab:sc2_1s2z}
\vspace{-20pt}
\end{table}

\textbf{Task Description.} In Section~\ref{sec:problem_statement}, we formulated the problem of modeling what we called the \emph{world state} $\fo$ of a dynamical system $\phi$ given \emph{local observations} $\{(\vx^{a}_{t}, \vO^{a}_{t})\}_{a=1}^{A}$ where $\vO_t^a = \phi(t, \fo)(\vx_t^a)$. Under certain restrictions, this problem can be mapped to that of multi-agent world modeling from partial and local observations, allowing us to evaluate the proposed model in a rich and challenging setting. In particular, we consider environments that are \textbf{(a)} \emph{spatial}, i.e. all agents $a$ in it have a well-defined and known location $\vx_t^a$ (at time $t$), \textbf{(b)} the agents' actions $\vu_t^a$ are local, in that their effects propagate away (from the agent) only at a finite speed, \textbf{(c)} the observations are local and centered around agents, in the sense that the agent only observes the events in its local vicinity, i.e., $\vO_t^a$. Observe that we do not fix the number of agents in the environment, and allow for agents to dynamically enter or exit the environment. Now, the task is: given observations $\vO_t^a$ from a team of (cooperating) agents at position $\vx_t^a$ and their corresponding actions $\vu_t^a$, predict the observation $\vO_{t'}^{q}$ that would be made by an agent at time $t' = t + 1$ if it were at position $\vx^q$. In particular, note that unlike in Bouncing Balls, the positions $\vx_t^a$ and $\vx_{t+1}^a$ are no longer independent and depend on the agents' behaviour. 

\textbf{The SC2 Domain.} Starcraft2 unit-micromanagement \citep{samvelyan2019starcraft} is a multi-agent reinforcement learning benchmark, wherein teams of heterogeneously typed units must defeat a team of opponents in melee and ranged combat. Each unit type has its own characteristics, e.g. maximum \emph{health}, \emph{shields}, weapon abilities (\emph{cool-down}, \emph{damage per second}, \emph{splash damage}, etc), and strengths (vulnerabilities) against (towards) other unit types, making the world-modeling task all the more rich and challenging. 

\textbf{Dataset.} The observations $\vO_t^a$ and actions $\vu_t^a$ are both multi-channel images represented in polar coordinates centered around the agent position $\vx_a^t$. The field of view (FOV) of each agent is therefore a circle of fixed radius centered around it. The channels of the image correspond to \textbf{(a)} a binary indicator marking whether a position in FOV is occupied by a living friendly agent (\emph{friendly marker}), \textbf{(b)} a categorical indicator marking the type of living units at a given position in FOV (\emph{unit-type marker}), and \textbf{(c)} four channels marking the health, energy, weapon-cooldown and shields (\emph{HECS markers}) of all agents in FOV. With a heuristic, we gather a total of $9$K trajectories $(\{\vx_t^a, \vO_t^a, \vu_t^a\}_{a=1}^{A})_{t = 1}^{100}$ spread over three training \emph{scenarios}, corresponding to \texttt{1c3s5z}\footnote{Here, the code $\texttt{1c3s5z}$ refers to a scenario where each team comprises $1$ \emph{colossus} (\texttt{1c}), $3$ \emph{stalkers} (\texttt{3s}), and $5$ \emph{zealots} (\texttt{5z}).}, \texttt{3s5z} and \texttt{2s5z} in \citet{samvelyan2019starcraft}. In addition, we also sample $1$K trajectories (each) from two OOD scenarios \texttt{1s2z} and \texttt{5s3z}. Details in Appendix~\ref{app:sc2}. 

\textbf{Training.} While adopting the training protocol detailed in Appendix~\ref{app:hparams_train}, we adapt the encoder and decoder architecture to match the state representation by including circular convolutions (cf. Appendix \ref{app:encdec_arch_sc2}). 
Now, recall that predicting the next state entails predicting images of binary friendly markers, categorical unit type markers and real valued HECS markers. Accordingly, the loss function is a sum of a binary cross-entropy term (on friendly markers), a categorical cross-entropy term (on unit-type markers) and a mean squared error term (on HECS markers). 

\textbf{Evaluation Criteria.} After having trained all models on scenarios \texttt{1c3s5z}, \texttt{3s5z} and \texttt{2s5z}, we test their robustness to dropped agents (Figure~\ref{fig:sc2_robustness}) and their performance on OOD scenarios (Table~\ref{tab:sc2_1s2z}). We only show baselines that achieve similar or better validation scores than S2RMs, and report the F1 scores for binary friendly markers, multi-class (macro) F1 score for unit-type markers, negative mean squared error for HECS markers (tables in Appendix~\ref{app:results_sc2}). 

\textbf{Results.} Figure~\ref{fig:sc2_robustness} shows that S2RMs remain robust when fewer agents supply their observations to the world model, whereas Table~\ref{tab:sc2_1s2z} shows that S2GRU outperforms the baselines in the OOD scenario \texttt{1s2z} but is matched by RMCs in \texttt{5s3z} (see Appendix~\ref{app:results_sc2} for details). 

\section*{Conclusions and Outlook}
We proposed Spatially Structured Recurrent Modules, a new class of models constructed to jointly leverage both spatial and modular structure in data, and explored its potential in the challenging problem setting of predicting the forward dynamics from partial observations at known spatial locations. In the tasks of video prediction from crops and multi-agent world modeling in the Starcraft2 domain, we found that it compares favorably against strong baselines in terms of out-of-distribution generalization and robustness to the number of available observations. Future work may focus on exploring efficient implementations using block-sparse methods \citep{gray2017gpu}, which could potentially unlock applications to significantly larger scale spatial problems encountered in domains such as humanitarian aid and climate change research \citep{rolnick2019tackling}. 

\section*{Acknowledgements}
The authors would like to thank Georgios Arvanitidis, Luigi Gresele for their feedback on the paper, and Murray Shanahan for the discussions. The authors also acknowledge the important role played by their colleagues at the Empirical Inference Department of MPI-IS T\"ubingen and Mila throughout the duration of this work.

\bibliography{example_paper}
\bibliographystyle{icml2019}

\clearpage
\newpage
\appendix
\begin{appendix}

\section{Starcraft2} \label{app:sc2}

\begin{figure*}[t]
\centering

\begin{subfigure}{0.48\textwidth}
\centering
\includegraphics[width=1\linewidth]{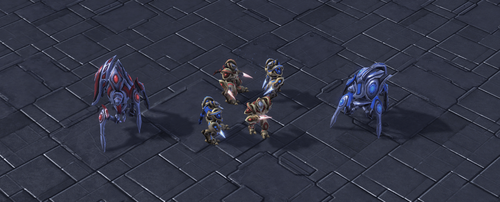}
\caption{\texttt{1s2z} (1 Stalker and 2 Zealots per team).}
\label{fig:sc2_ood_1s2z}
\end{subfigure}\hfill
\begin{subfigure}{0.48\textwidth}
\centering
\includegraphics[width=1\linewidth]{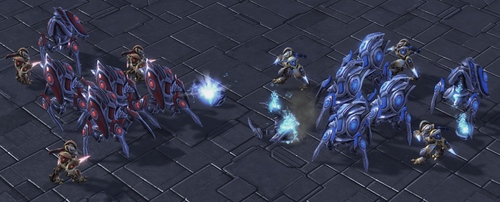}
\caption{\texttt{5s3z} (5 Stalkers and 3 Zealots per team).}
\label{fig:sc2_ood_5s3z}
\end{subfigure}

\medskip

\begin{subfigure}{0.48\textwidth}
\centering
\includegraphics[width=1\linewidth]{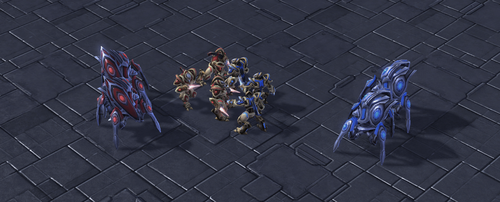}
\caption{\texttt{2s3z} (2 Stalkers and 3 Zealots per team).}
\label{fig:sc2_id_2s3z}
\end{subfigure}\hfill
\begin{subfigure}{0.48\textwidth}
\centering
\includegraphics[width=1\linewidth]{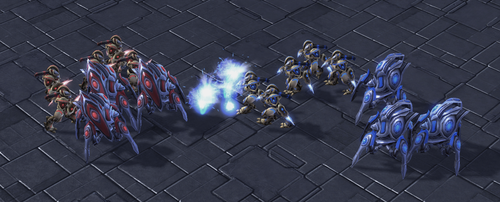}
\caption{\texttt{3s5z} (3 Stalkers and 5 Zealots per team).}
\label{fig:sc2_id_3s5z}
\end{subfigure}

\medskip

\begin{subfigure}{0.48\textwidth}
\centering
\includegraphics[width=1\linewidth]{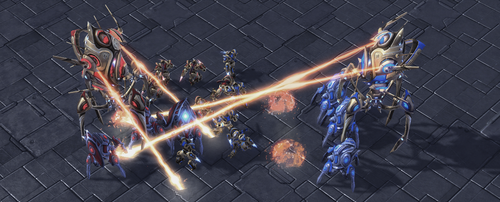}
\caption{\texttt{1c3s5z} (1 Colossus, 3 Stalkers and 5 Zealots per team).}
\label{fig:sc2_id_1c3s5z}
\end{subfigure}

\caption{Human readable illustrations of the Starcraft2 (SMAC) scenarios we consider in this work. Figures~\ref{fig:sc2_ood_1s2z} and \ref{fig:sc2_ood_5s3z} show the OOD scenarios, whereas Figures~\ref{fig:sc2_id_2s3z}, \ref{fig:sc2_id_3s5z} and \ref{fig:sc2_id_1c3s5z} show the training scenarios (provided by \citet{samvelyan2019starcraft}).}
\label{fig:sc2_scenarios}

\end{figure*}

The Starcraft2 Environment we use is a modified version of the SMAC-Env proposed in \citet{samvelyan2019starcraft} and built on PySC2 wrapper around Blizzard SC2 API \citep{vinyals2017starcraft}. Starcraft2 is a real-time-strategy (RTS) game where players are tasked with manufacturing and controlling armies of \emph{units} (airborne or land-based) to defeat the opponent's army (where the opponent can be an AI or another human). The players must choose their \emph{alien race}\footnote{Please note that this is a game-specific notion.} before starting the game;  available options are \emph{Protoss}, \emph{Terran} and \emph{Zerg}. All unit types (of all races) have their strengths and weaknesses against other unit types, be it in terms of maximum health, shields (Protoss), energy (Terran), DPS (damage per second, related to weapon cooldown), splash damage, or manufacturing costs (measured in \emph{minerals} and \emph{vespene gas}, which must be mined). 

The key engineering contribution of \citet{samvelyan2019starcraft} is to repurpose the RTS game as a multi-agent environment, where the individual units in the army become individual agents\footnote{Note that this is rather unconventional, since each player usually controls entire armies and must switch between macro- and micro-management of units or unit-groups.}. The result is a rich and challenging environment where heterogeneous teams of agents must defeat each other in melee and ranged combat. The composition of teams vary between \emph{scenarios}, of which \citet{samvelyan2019starcraft} provide a selection. Further, new scenarios can be easily created with the SC2MapEditor, which allows for practically endlessly many possibilities. 

We build on \citet{samvelyan2019starcraft} by modifying their environment to better expose the transfer and out-of-distribution aspects of the domain by (a) standardizing the state and action space across a large class of scenarios and (b) standardizing the unit stats to better reflect the game-defined notion of hit-points. 

\subsection{Standardized State Space for All Scenarios}
In the environment provided by \citet{samvelyan2019starcraft}, the dimensionality of the vector state space varies with the number of friendly and enemy agents, which in turn varies with the scenario. While this is not an issue in the typical use case of training MARL agents in a fixed scenario, it is not convenient for designing models that seamlessly handle multiple scenarios. In the following, we propose an alternate state representation that preserves the spatial structure and is consistent across multiple scenarios. 

Instead of representing the state of an agent $a$ with a vector of variable dimension, we represent it with a multi-channel \emph{polar image} $\fI^a$ of shape $C \times I \times J$, where $C$ is the number of channels and $(I, J)$ is the image size. Given the \emph{radial} and \emph{angular} resolutions $\rho$ and $\varphi$ (respectively), the pixel coordinate $i = 0, ..., I-1, j = 0, ..., J-1$ corresponds to coordinates $(i \cdot \rho, j \cdot \varphi)$ with respect to a polar coordinate system centered on the agent $a$, where the positive $x$-axis ($j = 0$) points towards the east. Further, the field of view (FOV) of an agent is characterized by a circle of radius $I\cdot \rho$ centered on the agent at 2D game-coordinates $\vx^a = (x_1^a, x_2^a)$, to which the Starcraft2 API \citep{vinyals2017starcraft} provides raw access. 

The polar image $\fI^a$ therefore provides an agent-centric view of the environment, where pixel coordinates $i, j$ in $\fI^a$ can be mapped to global game coordinates $\vx = (x_1, x_2)$ in FOV via: 
\begin{align}
x_1 &= i \cdot \rho \cos{\left[j \cdot \varphi\right]} + x_1^a \\
x_2 &= i \cdot \rho \sin{\left[j \cdot \varphi\right]} + x_2^a
\end{align}
In what follows, we denote this transformation with $T_a$, as in $T_a(i, j) = (x_1, x_2)$.

Now, the channels in the polar image can encode various aspects of the observation; in our case: friendly markers (one channel), unit-type markers (nine channels, one-hot), health-energy-cooldown-shields (HECS, four channels) and terrain height (one channel).  As an example, let us consider the friendly markers, which is a binary indicator marking units that are friendly. If we have an agent at game position $(x_1, x_2)$ that is friendly to agent $a$, then we would expect the pixel coordinate $(i, j) = T_a^{-1}(x_1, x_2)$ of the corresponding channel in the polar image $\fI^a$ to be $1$, but $0$ otherwise. Likewise, the value of $\fI$ at the channels corresponding to HECS at pixel position $i, j$ gives the HECS of the corresponding unit\footnote{If health drops to zero, the unit is considered dead and the representation does not differentiate between dead and absent units.} at $T_a(i, j)$. 
This representation has the following advantages: \textbf{(a)} it does not depend on the number of units in the field of view, \textbf{(b)} it exposes the spatial structure in the arrangement of units which can naturally processed by convolutional neural networks (e.g. with circular convolutions). 

Nevertheless, it has the disadvantage that the positions are \emph{quantized} to pixels, but the euclidean distance between the locations represented by pixels $(i, j)$ and $(i, j + 1)$ increases with increasing $i$. Consequently, this representation may not remain suitable for larger FOVs. 

Further, this representation is also appropriate for the action space. Given an agent, we represent the one-hot categorical actions of all friendly agents in FOV as a multi-channel polar image. In this representation, the pixel position $i, j$ gives the action taken by an agent at at position $T_{a}(i, j)$. Unfriendly agents get assigned an "unknown action", whereas positions not occupied by a living agent are assigned a "no-op" action. 

\subsection{Standardized Unit Stats} 
At any given point in time, an active unit in Starcraft2 has certain \emph{stats}, e.g. its health, energy (Terran), shields (Protoss) and weapon-cooldown (for armed units). A large and expensive unit-type like the Colossus has more max-health (\emph{hit-points}) than smaller units like Stalkers and Marines\footnote{These stats may change with game-versions, and are catalogued here: \url{https://liquipedia.net/starcraft2/Units_(StarCraft)}.}. Likewise, unit-types differ in the rate at which they deal damage (measured in damage-per-second or DPS, excluding splash damage), which in turn depends on the cooldown duration of the active weapon. 

Now, the environment provided by \citet{samvelyan2019starcraft} normalizes the stats by their respective maximum value, resulting in values between $0$ and $1$. However, given that different units may have different normalization, the stats are rendered incomparable between unit types (without additionally accounting the unit-type). We address this by standardizing stats (instead of normalizing) by dividing them by a fixed value. In this scheme, the stats are scaled uniformly across all unit-types, enabling models to directly rely on them instead of having to account for the respective unit-types. 

\section{Hyperparameters and Architectures} \label{app:hparams_and_archs}
\subsection{Encoder and Decoder for Bouncing Balls} \label{app:encdec_arch_bb}
The architectures of image encoder and decoder was fixed for all models after initial experimentation. We converged to the following architectures. 

\begin{figure}
\centering
\begin{subfigure}{.24\textwidth}
  \centering
  \includegraphics[width=1\linewidth]{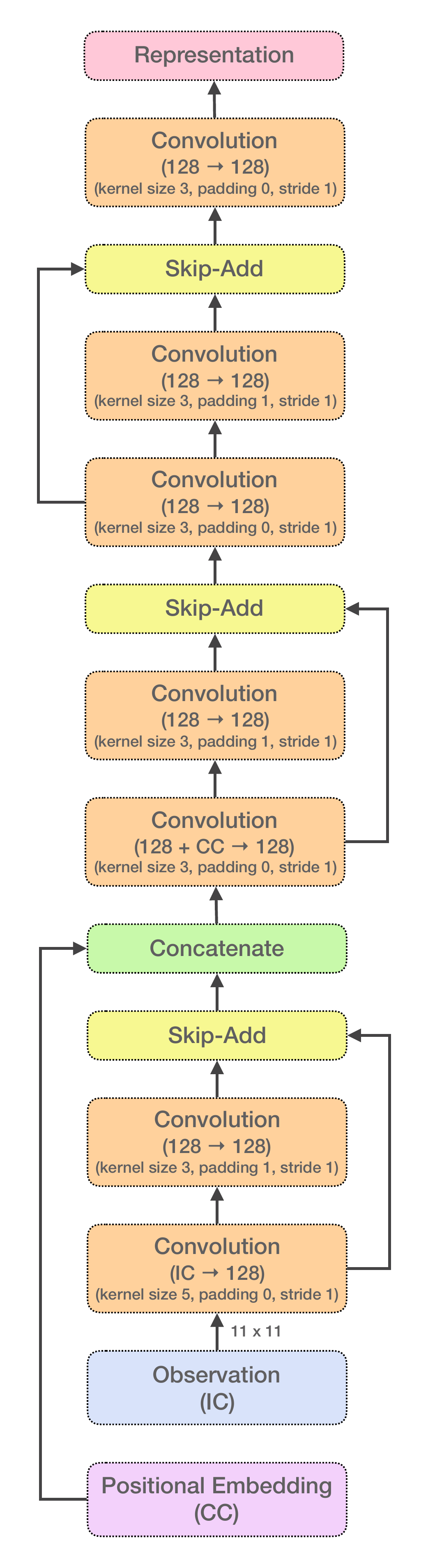}
  \caption{Encoder.}
  \label{fig:sub1}
\end{subfigure}%
\begin{subfigure}{.24\textwidth}
  \centering
  \includegraphics[width=1\linewidth]{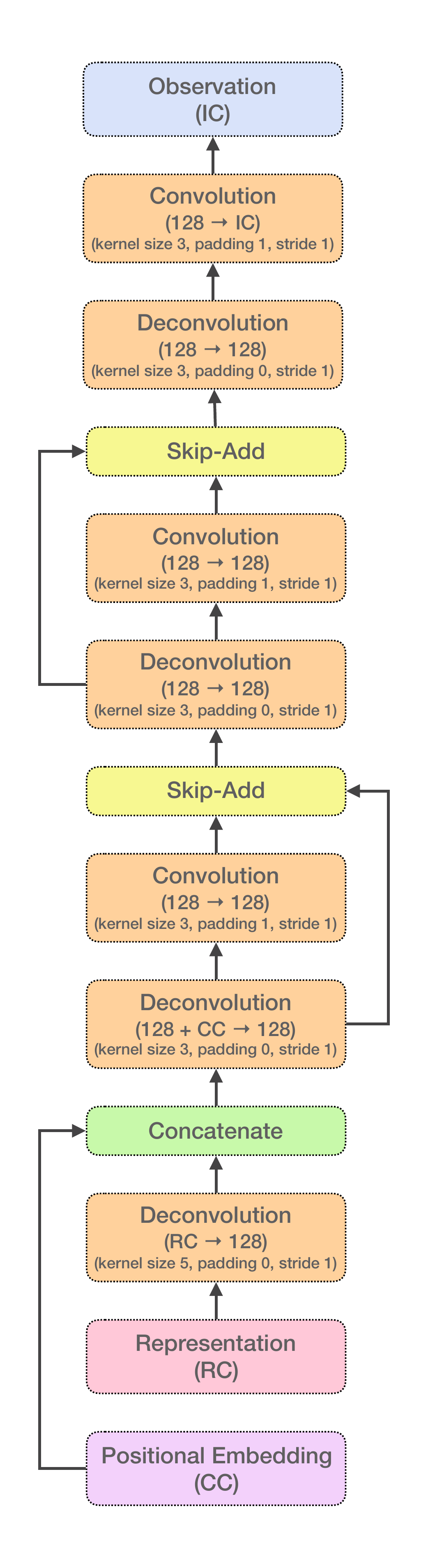}
  \caption{Decoder.}
  \label{fig:sub2}
\end{subfigure}
\caption{Baseline encoder and decoder architectures for the Bouncing Ball task.}
\label{fig:bb_baseline_enc_dec_arch}
\end{figure}

\begin{figure}
\centering
\begin{subfigure}{.24\textwidth}
  \centering
  \includegraphics[width=1\linewidth]{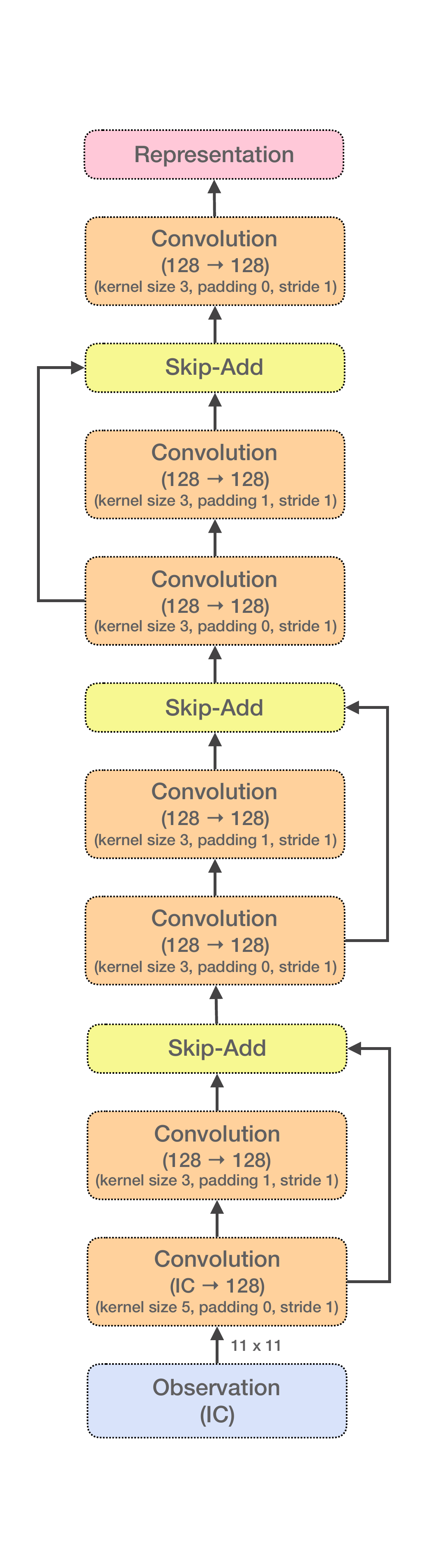}
  \caption{Encoder.}
  \label{fig:sub1}
\end{subfigure}%
\begin{subfigure}{.24\textwidth}
  \centering
  \includegraphics[width=1\linewidth]{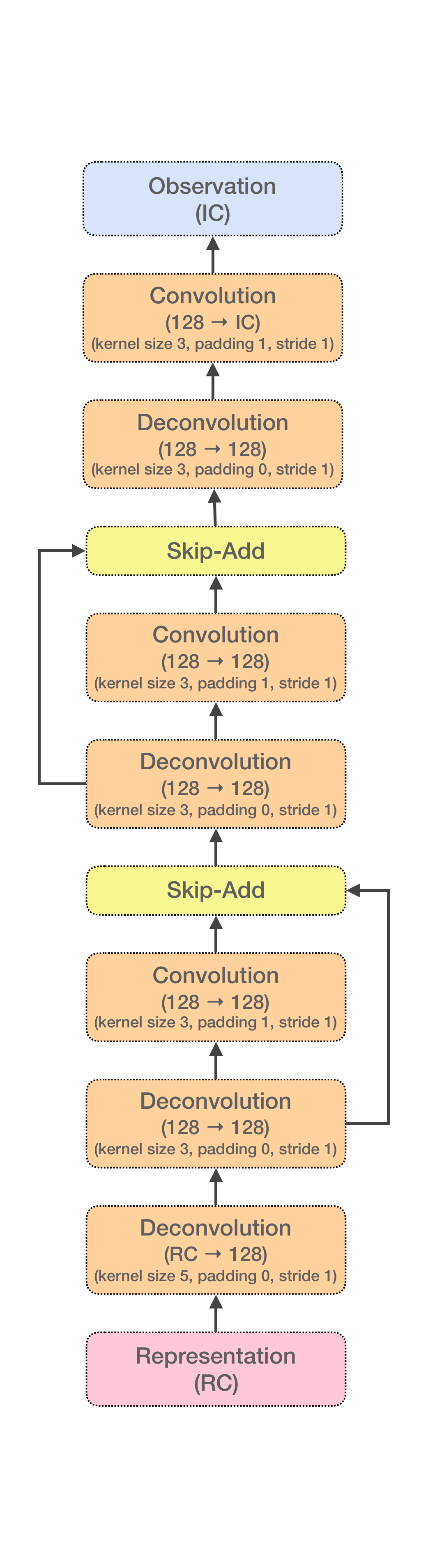}
  \caption{Decoder.}
  \label{fig:sub2}
\end{subfigure}
\caption{S2RM encoder and decoder architectures for the Bouncing Ball task.}
\label{fig:bb_s2rm_enc_dec_arch}
\end{figure}

\subsubsection{S2RMs}
The encoder (decoder) is a (de)convolutional network with residual connections (Figure~\ref{fig:bb_s2rm_enc_dec_arch}). 

\subsubsection{Baselines}
Like in the case of S2RMs, the encoder (decoder) is a (de)convolutional network with residual connections (Figure~\ref{fig:bb_baseline_enc_dec_arch}), but with the positional embeddings injected after the second convolutional layer. This is loosely inspired by the encoders used in \citet{eslami2018neural}. 

\subsection{Encoder and Decoder for Starcraft2} \label{app:encdec_arch_sc2}

\begin{figure}
\centering
\begin{subfigure}{.24\textwidth}
  \centering
  \includegraphics[width=1\linewidth]{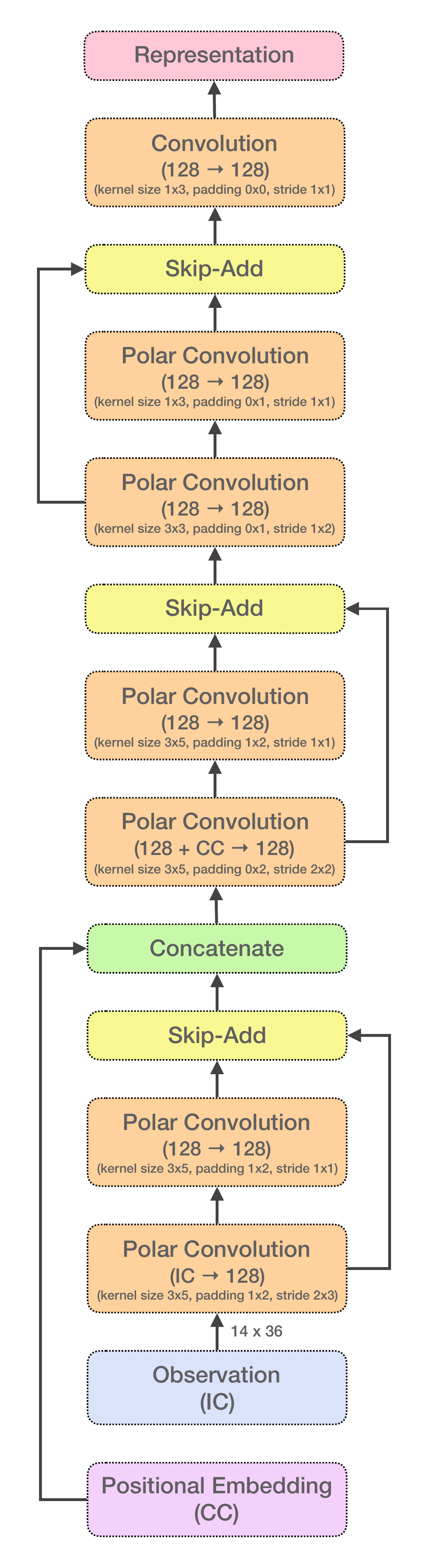}
  \caption{Encoder.}
  \label{fig:sub1}
\end{subfigure}%
\begin{subfigure}{.24\textwidth}
  \centering
  \includegraphics[width=1\linewidth]{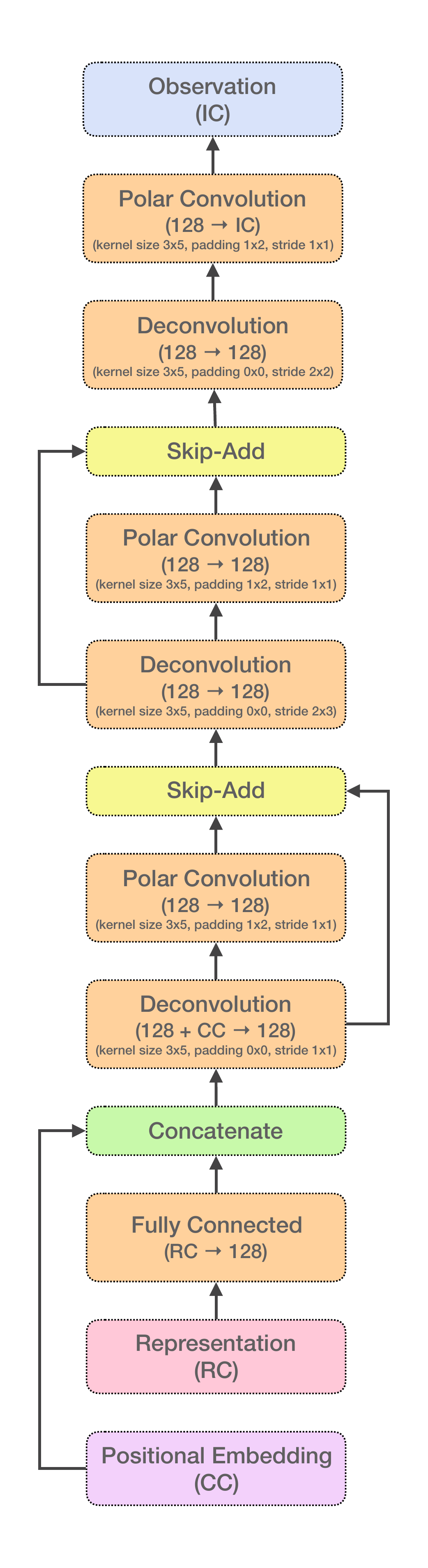}
  \caption{Decoder.}
  \label{fig:sub2}
\end{subfigure}
\caption{Baseline encoder and decoder architectures for the Starcraft2 task.}
\label{fig:sc2_baseline_enc_dec_arch}
\end{figure}

\begin{figure}
\centering
\begin{subfigure}{.24\textwidth}
  \centering
  \includegraphics[width=1\linewidth]{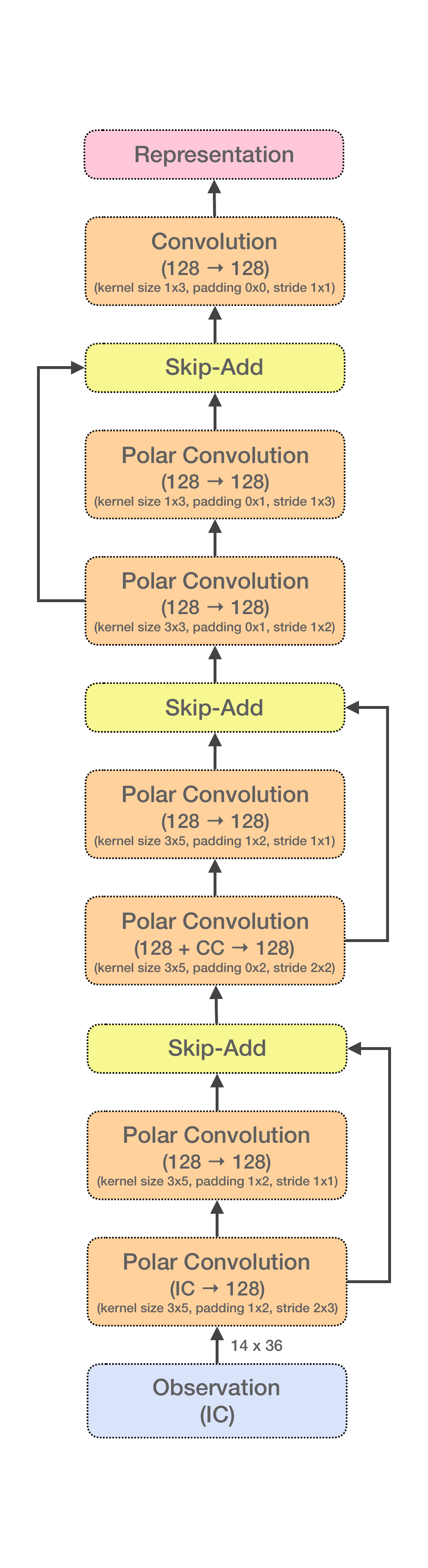}
  \caption{Encoder.}
  \label{fig:sub1}
\end{subfigure}%
\begin{subfigure}{.24\textwidth}
  \centering
  \includegraphics[width=1\linewidth]{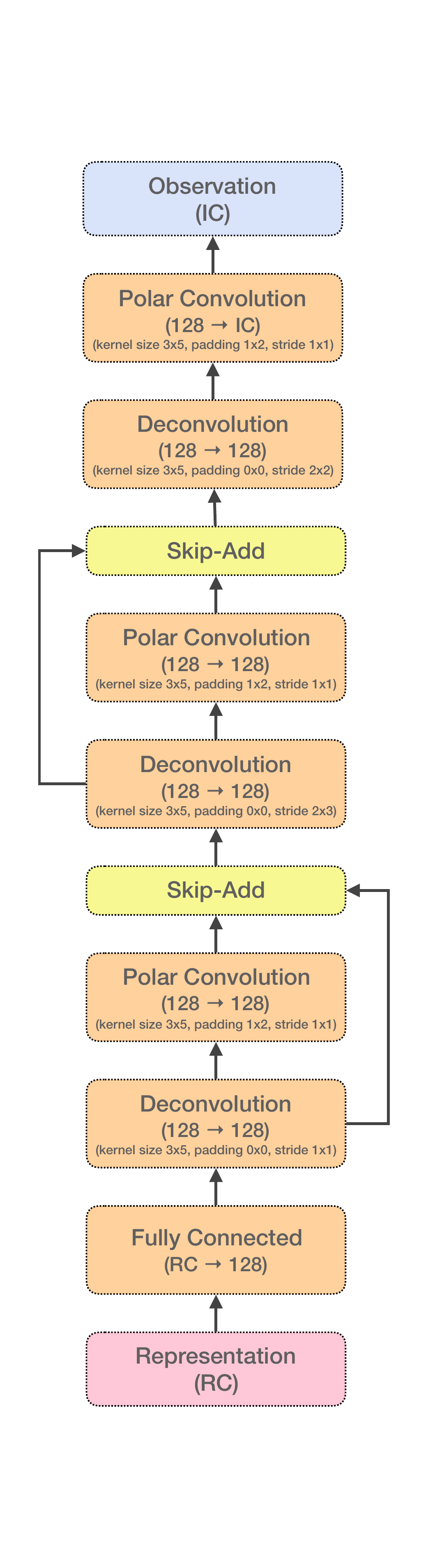}
  \caption{Decoder.}
  \label{fig:sub2}
\end{subfigure}
\caption{S2RM encoder and decoder architectures for the Starcraft2 task.}
\label{fig:sc2_s2rm_enc_dec_arch}
\end{figure}

\subsubsection{S2RMs}
Recall from Appendix~\ref{app:sc2} that the states are polar images. We therefore use \emph{polar convolutions}, which entails zero-padding the input image along the first (\emph{radial}) dimension but circular padding along the second (\emph{angular}) dimension. The encoder and decoder architectures can be found in Figure~\ref{fig:sc2_s2rm_enc_dec_arch}. 

\subsubsection{Baselines}
Like for S2RMs, we use polar convolutions while injecting the positional embeddings further downstream in the network. The corresponding encoder and decoder architectures are illustrated in Figure~\ref{fig:sc2_baseline_enc_dec_arch}. 

\subsection{Spatially Structured Relational Memory Cores (S2RMCs)} \label{app:s2rmc}

Embedding Relational Memory Cores \citep{santoro2018relational} na\"ively in the S2RM architecture did not result in a working model. We therefore had to adapt it by first projecting the memory matrix ($M$ in \citet{santoro2018relational}) of the $m$-th RMC to a \emph{message} $\vh_{t}^m$. This message is then processed by the intercell attention to obtain $\bar \vh_t^m$, which is finally concatenated with the memory matrix and current input before applying the attention mechanism (i.e. in Equation~2 of \citet{santoro2018relational}, we replace $\left[M; x\right]$ with $\left[M; x, \bar \vh_t^m \right]$). 

\subsection{Hyperparameters} \label{app:hparams}
\subsubsection{Bouncing Ball Models} \label{app:hparams_bb}

The hyperparameters we used can be found in Table~\ref{tab:bb_hparams}. Further, note that in Equation~\ref{eq:zonalkernel}, we pass the gradients through the constant region of the kernel as if the kernel had not been truncated. 

\begin{table}[]
    \centering
    \begin{tabular}{lr}
    \toprule
    \textbf{Model}                 &         \\
    Hyperparameter                 &   Value \\
    \midrule
    \textbf{S2GRU}                 &      \\
    Number of modules ($M$)        & 10   \\
    GRU: hidden size per module    & 128  \\
    Module embedding size ($d$)    & 16   \\
    Kernel bandwidth ($\epsilon$)  & 1    \\
    Kernel truncation ($\tau$)     & 0.6  \\
    \texttt{shape $\Theta^{(Q/K)}$} & (128, 2, 016) \\
    \texttt{shape $\Theta^{(V)}$}   & (128, 2, 128) \\
    \texttt{shape $\Phi^{(Q/K)}$}   & (128, 4, 016) \\
    \texttt{shape $\Phi^{(V)}$}     & (128, 4, 128) \\
    \midrule
    \textbf{RMC} \citep{santoro2018relational}  &     \\
    Number of attention heads      & 4   \\
    Size of attention head         & 128 \\
    Number of memory slots         & 1   \\
    Key size                       & 128 \\
    \midrule
    \textbf{LSTM} \citep{hochreiter1997long} &      \\
    Hidden size                   & 512  \\
    \midrule
    \textbf{RIMs} \citep{goyal2019recurrent}  &     \\
    Number of RIMs ($k_T$)        & 6   \\
    Update Top-k ($k_A$)          & 5   \\
    Hidden size ($h_{size}$)      & 510 \\
    Input key size                & 32  \\
    Input value size              & 400 \\
    \midrule
    \textbf{TTO}                  &     \\
    MLP hidden size               & 512 \\
    \bottomrule
    \end{tabular}
    \caption{Hyperparameters used for various models on the Bouncing Ball task. Hyperparameters not listed here were left at their respective default values.}
    \label{tab:bb_hparams}
\end{table}

\subsubsection{Starcraft2 Models} \label{app:hparams_sc2}

The hyperparameters we used can be found in Table~\ref{tab:sc2_hparams}. Note that we only report models that attained a validation loss similar to or better than S2RMs. 

\begin{table}[]
    \centering
    \begin{tabular}{lr}
    \toprule
    \textbf{Model}                 &       \\
    Hyperparameter                 &   Value \\
    \midrule
    \textbf{S2GRUs}                &      \\
    Number of modules ($M$)        & 10   \\
    GRU: hidden size per module    & 128  \\
    Module embedding size ($d$)    & 8    \\
    Kernel bandwidth ($\epsilon$)  & 1    \\
    Kernel truncation ($\tau$)     & 0.5  \\
    \texttt{shape $\Theta^{(Q/K)}$} & (128, 2, 016) \\
    \texttt{shape $\Theta^{(V)}$}   & (128, 2, 128) \\
    \texttt{shape $\Phi^{(Q/K)}$}   & (128, 4, 016) \\
    \texttt{shape $\Phi^{(V)}$}     & (128, 4, 128) \\
    \midrule
    \textbf{S2RMC}                 &      \\
    Number of modules ($M$)        & 10   \\
    RMC: number of attention heads & 4    \\
    RMC: size of attention head    & 64   \\
    RMC: number of memory slots    & 4    \\
    RMC: key size                  & 64   \\
    Module embedding size ($d$)    & 8    \\
    Kernel bandwidth ($\epsilon$)  & 1    \\
    Kernel truncation ($\tau$)     & 0.5  \\
    \texttt{shape $\Theta^{(Q/K)}$} & (128, 2, 016) \\
    \texttt{shape $\Theta^{(V)}$}   & (128, 2, 128) \\
    \texttt{shape $\Phi^{(Q/K)}$}   & (128, 4, 016) \\
    \texttt{shape $\Phi^{(V)}$}     & (128, 4, 128) \\
    \midrule
    \textbf{RMC} \citep{santoro2018relational}  &     \\
    Number of attention heads      & 4   \\
    Size of attention head         & 128 \\
    Number of memory slots         & 1   \\
    Key size                       & 16  \\
    \midrule
    \textbf{LSTM} \citep{hochreiter1997long} &      \\
    Hidden size                   & 2048 \\
    \midrule
    \textbf{TTO}                  &     \\
    MLP hidden size               & 512 \\
    \bottomrule
    \end{tabular}
    \caption{Hyperparameters used for various models on the Starcraft2 task. Hyperparameters not listed here were left at their respective default values.}
    \label{tab:sc2_hparams}
\end{table}

\subsubsection{Training} \label{app:hparams_train}
All models were trained using Adam \cite{kingma2014adam} with an initial learning rate $0.0003$\footnote{\url{https://twitter.com/karpathy/status/801621764144971776?s=20}}. We use Pytorch's \citep{paszke2019pytorch} \texttt{ReduceLROnPlateau} learning rate scheduler to decay the learning rate by a factor of $2$ if the validation loss does not improve by at least $0.01\%$ over the span of $5$ epochs. We initially train all models for $100$ epochs, select the best of three successful runs, fine-tune it for another $100$ epochs, and finally select the checkpoint with the lowest validation loss (i.e. we early stop). We train all models with batch-size 8 (Starcraft2) or 32 (Bouncing Balls) on a single V100-32GB GPU (each). 

\section{Additional Results} \label{app:results}
\subsection{Bouncing Balls} \label{app:results_bb}

\subsubsection{Rollouts}
To obtain the rollouts in Figure~\ref{fig:bb_roll}, we adopt the following strategy. For the first $20$ \emph{prompt-steps}, we present all models with exactly the same $11 \times 11$ crops around randomly sampled pixel positions for $20$ time-steps. For the next $25$ steps, all models are queried at random pixel positions\footnote{These random pixel positions are the same for all models, but change between time-steps}, and the resulting predictions (on crops) are thresholded at $0.5$ and fed back in to the model for the next step (at known pixel positions from the previous step). 

\begin{figure*} 
\centering
\includegraphics[width=1\textwidth]{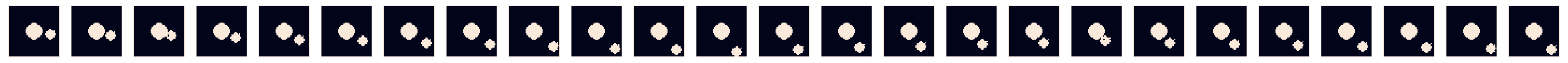}
\includegraphics[width=1\textwidth]{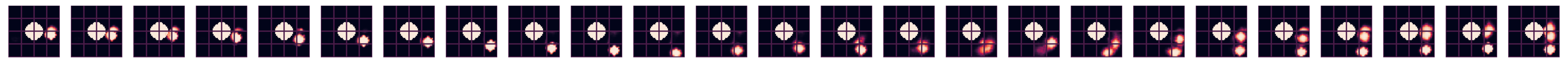}
\includegraphics[width=1\textwidth]{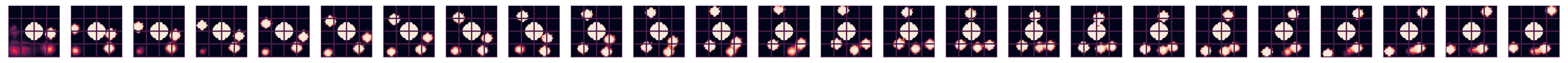}
\includegraphics[width=1\textwidth]{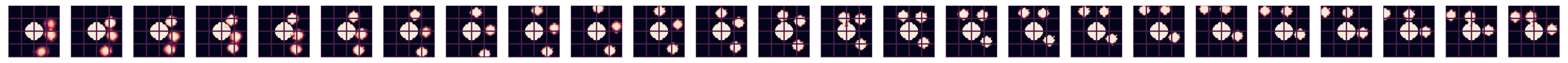}
\includegraphics[width=1\textwidth]{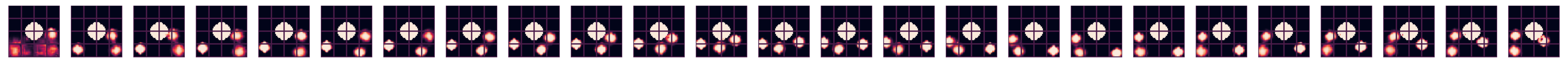}
\caption{Rollouts (OOD) with 1 bouncing ball, from top to bottom: ground-truth, S2GRU, RIMs, RMC, LSTM. Note that all models were trained on sequences with 3 bouncing balls, and the global state was reconstructed by stitching together $11 \times 11$ patches from the models (queried on a $4 \times 4$ grid).}
\label{fig:bb_roll_1b}
\end{figure*}

\begin{figure*} 
\centering
\includegraphics[width=1\textwidth]{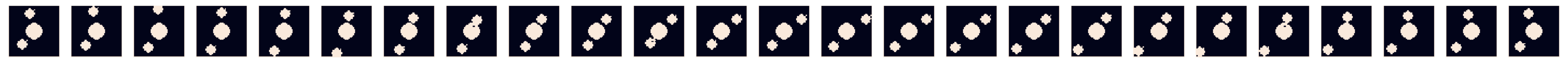}
\includegraphics[width=1\textwidth]{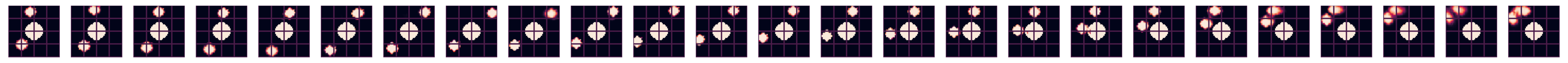}
\includegraphics[width=1\textwidth]{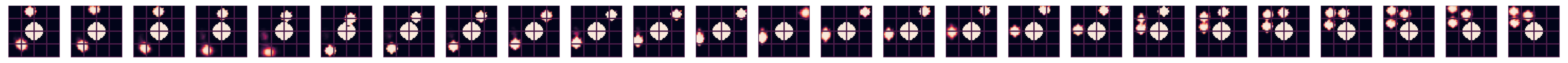}
\includegraphics[width=1\textwidth]{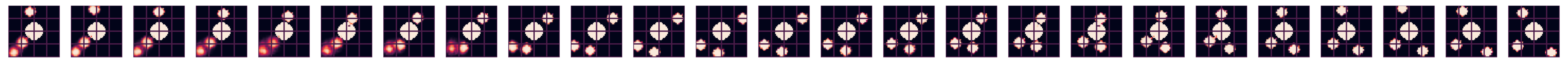}
\includegraphics[width=1\textwidth]{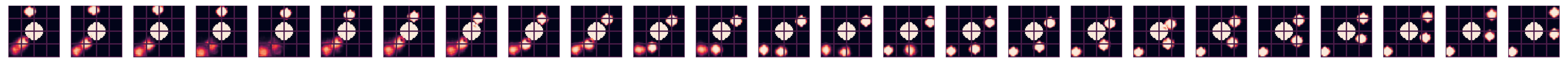}
\caption{Rollouts (OOD) with 2 bouncing balls, from top to bottom: ground-truth, S2GRU, RIMs, RMC, LSTM. Note that all models were trained on sequences with 3 bouncing balls, and the global state was reconstructed by stitching together $11 \times 11$ patches from the models (queried on a $4 \times 4$ grid).}
\label{fig:bb_roll_2b}
\end{figure*}

\begin{figure*} 
\centering
\includegraphics[width=1\textwidth]{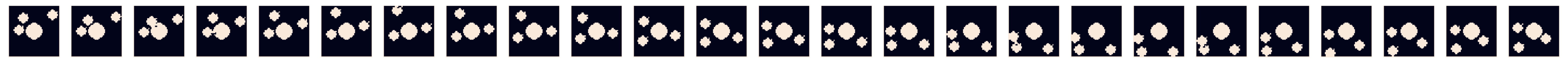}
\includegraphics[width=1\textwidth]{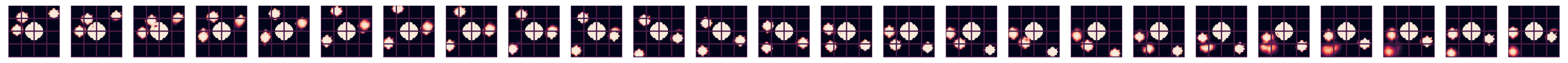}
\includegraphics[width=1\textwidth]{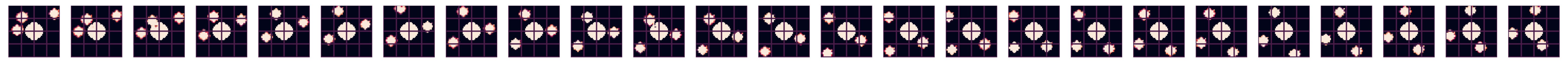}
\includegraphics[width=1\textwidth]{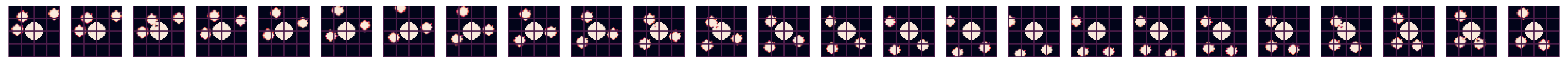}
\includegraphics[width=1\textwidth]{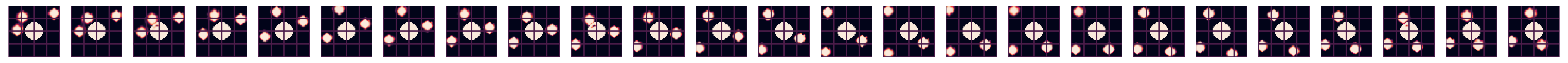}
\caption{Rollouts (ID) with 3 bouncing balls, from top to bottom: ground-truth, S2GRU, RIMs, RMC, LSTM. Note that all models were trained on sequences with 3 bouncing balls, and the global state was reconstructed by stitching together $11 \times 11$ patches from the models (queried on a $4 \times 4$ grid).}
\label{fig:bb_roll_3b}
\end{figure*}

\begin{figure*} 
\centering
\includegraphics[width=1\textwidth]{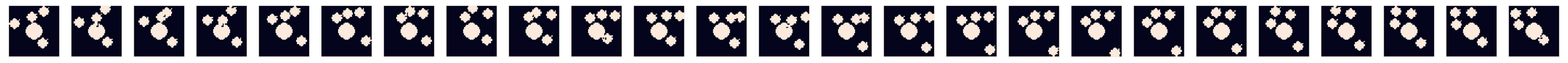}
\includegraphics[width=1\textwidth]{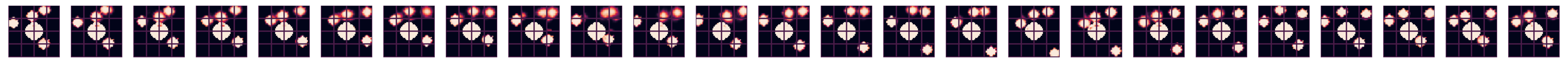}
\includegraphics[width=1\textwidth]{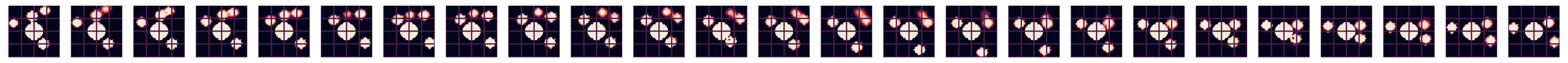}
\includegraphics[width=1\textwidth]{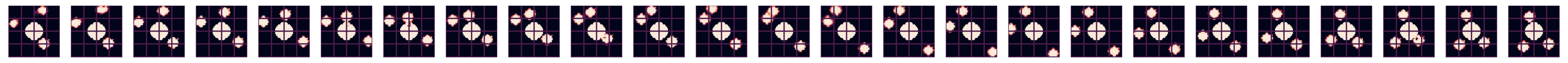}
\includegraphics[width=1\textwidth]{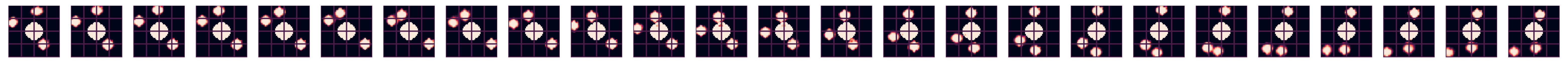}
\caption{Rollouts (OOD) with 4 bouncing balls, from top to bottom: ground-truth, S2GRU, RIMs, RMC, LSTM. Note that all models were trained on sequences with 3 bouncing balls, and the global state was reconstructed by stitching together $11 \times 11$ patches from the models (queried on a $4 \times 4$ grid).}
\label{fig:bb_roll_4b}
\end{figure*}

\begin{figure*} 
\centering
\includegraphics[width=1\textwidth]{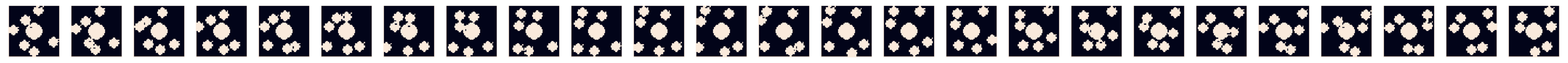}
\includegraphics[width=1\textwidth]{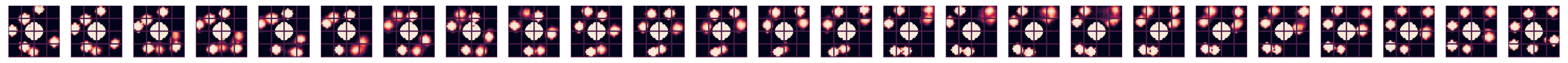}
\includegraphics[width=1\textwidth]{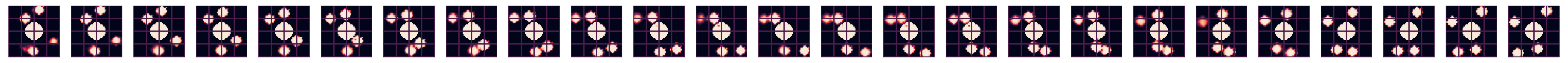}
\includegraphics[width=1\textwidth]{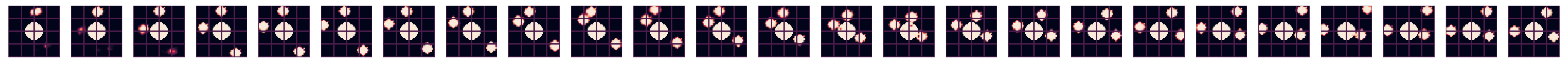}
\includegraphics[width=1\textwidth]{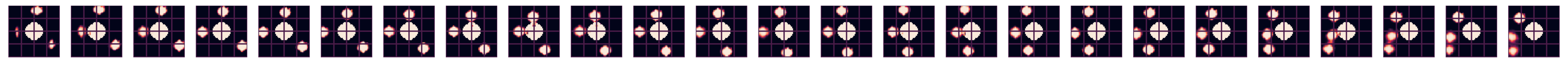}
\caption{Rollouts (OOD) with 6 bouncing balls, from top to bottom: ground-truth, S2GRU, RIMs, RMC, LSTM. Note that all models were trained on sequences with 3 bouncing balls, and the global state was reconstructed by stitching together $11 \times 11$ patches from the models (queried on a $4 \times 4$ grid).}
\label{fig:bb_roll_6b}
\end{figure*}

Also at every time-step, the models are queried for their predictions on $16$ pixel locations placed on a $4 \times 4$ grid. The resulting predictions are stitched together and shown in Figures~\ref{fig:bb_roll_1b}, \ref{fig:bb_roll_2b}, \ref{fig:bb_roll_3b}, \ref{fig:bb_roll_4b}, \ref{fig:bb_roll} and \ref{fig:bb_roll_6b}. 

\subsubsection{Robustness to Dropped Views}

\begin{figure*}[htp]
\centering

\begin{subfigure}{0.5\textwidth}
\centering
\includegraphics[width=\textwidth]{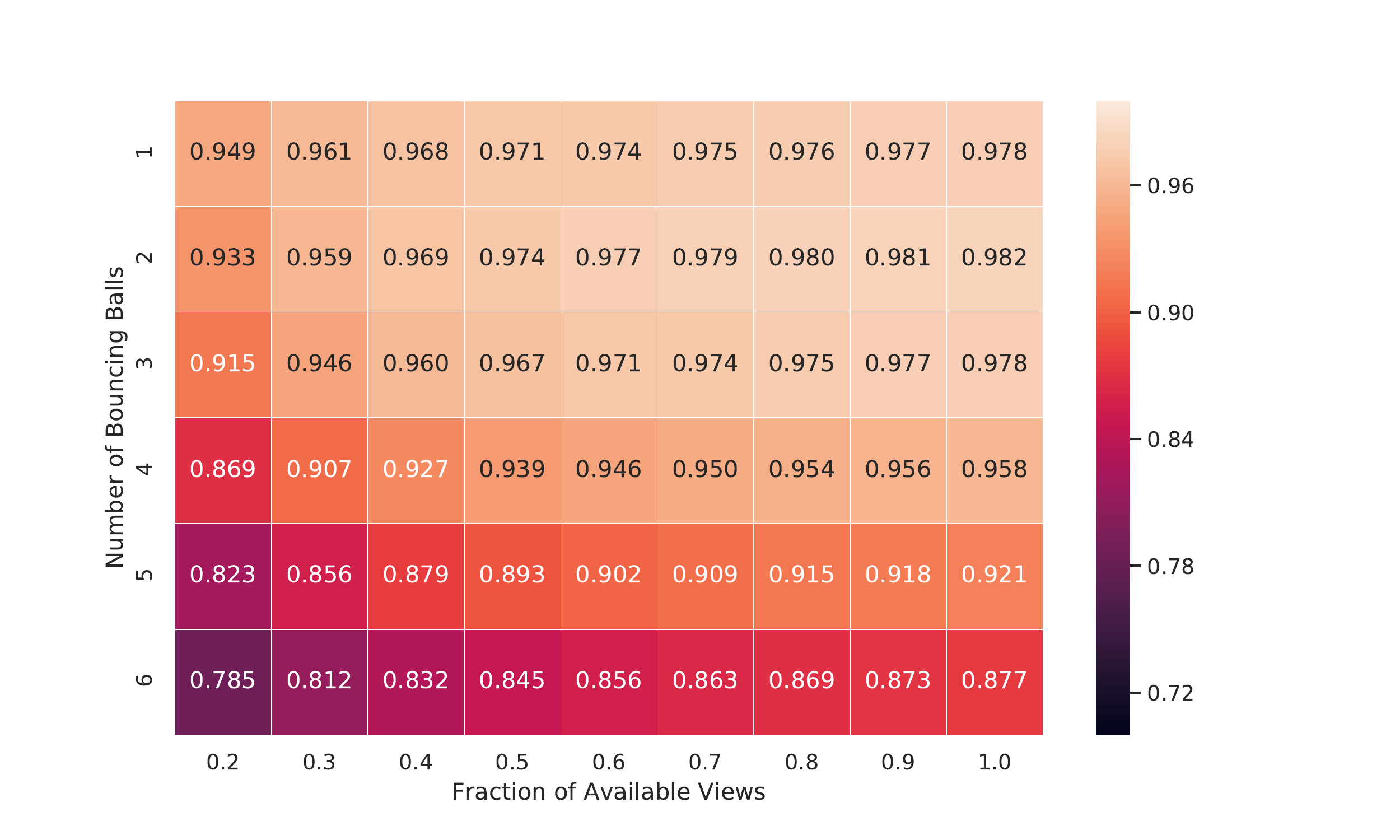}
\caption{S2GRU}
\label{fig:heatmaps_bacc_s2gru}
\end{subfigure}\hfill
\begin{subfigure}{0.5\textwidth}
\centering
\includegraphics[width=\textwidth]{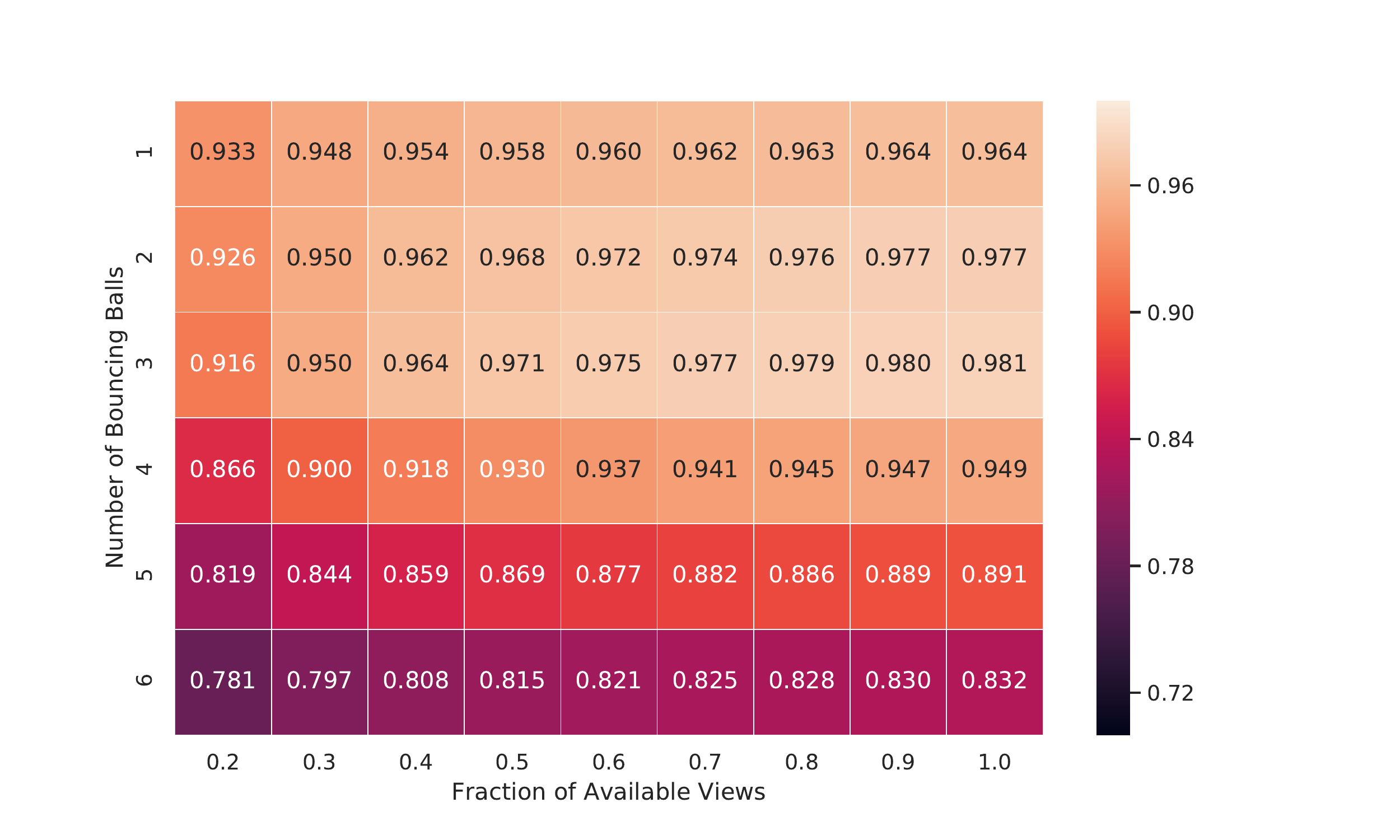}
\caption{RIMs}
\label{fig:heatmaps_bacc_rims}
\end{subfigure}

\medskip

\begin{subfigure}{0.5\textwidth}
\centering
\includegraphics[width=\textwidth]{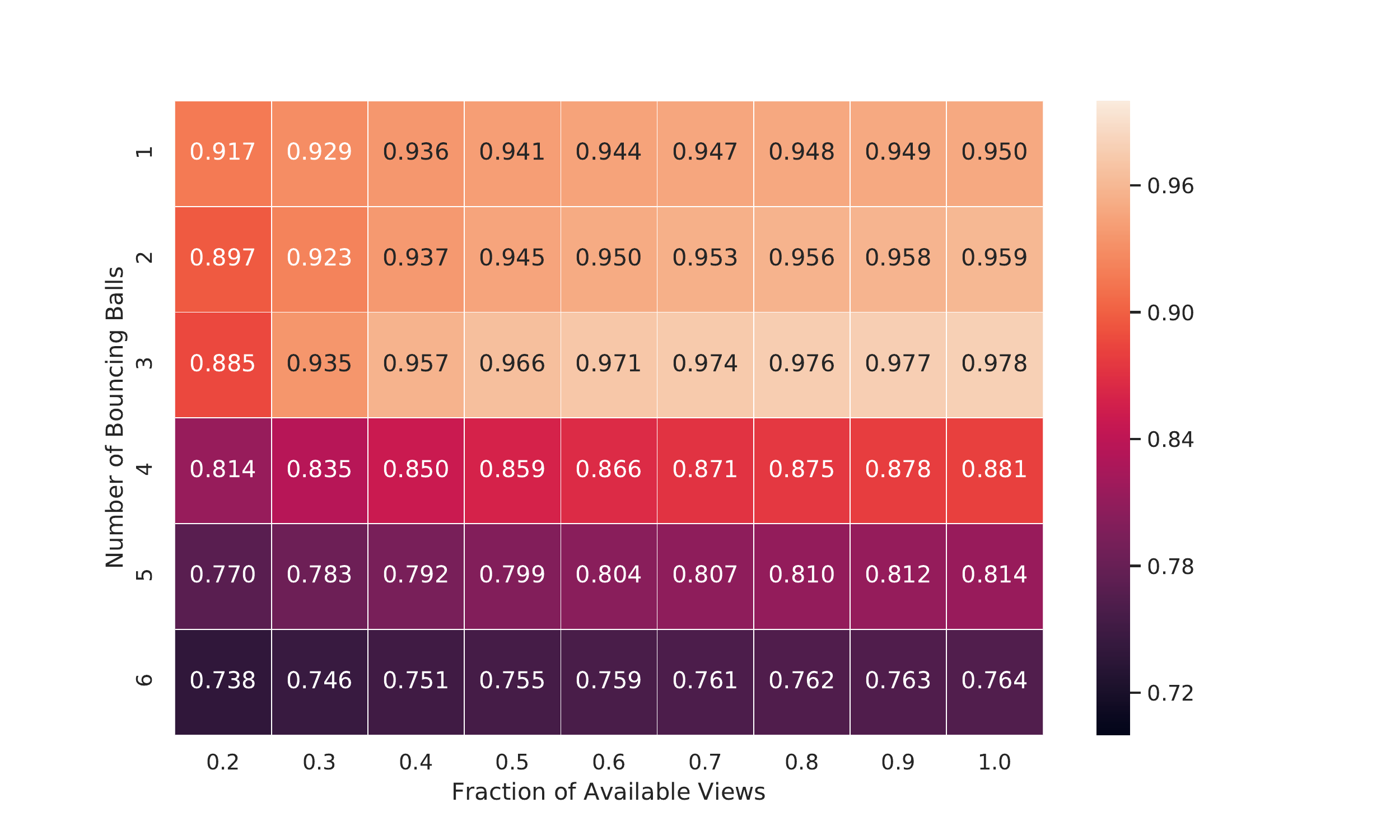}
\caption{RMC}
\label{fig:heatmaps_bacc_rmc}
\end{subfigure}\hfill
\begin{subfigure}{0.5\textwidth}
\centering
\includegraphics[width=\textwidth]{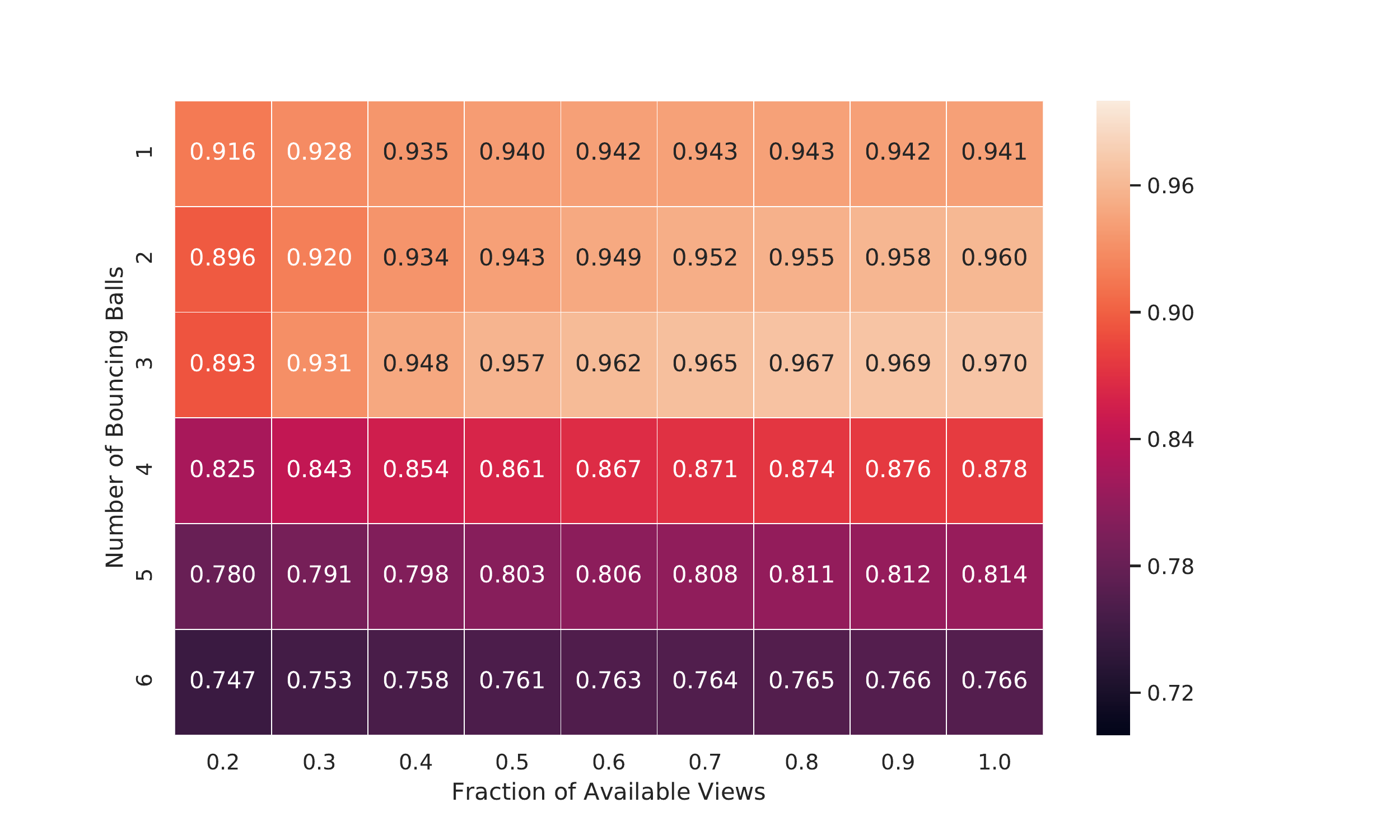}
\caption{LSTM}
\label{fig:heatmaps_bacc_lstm}
\end{subfigure}

\medskip

\begin{subfigure}{0.5\textwidth}
\centering
\includegraphics[width=\textwidth]{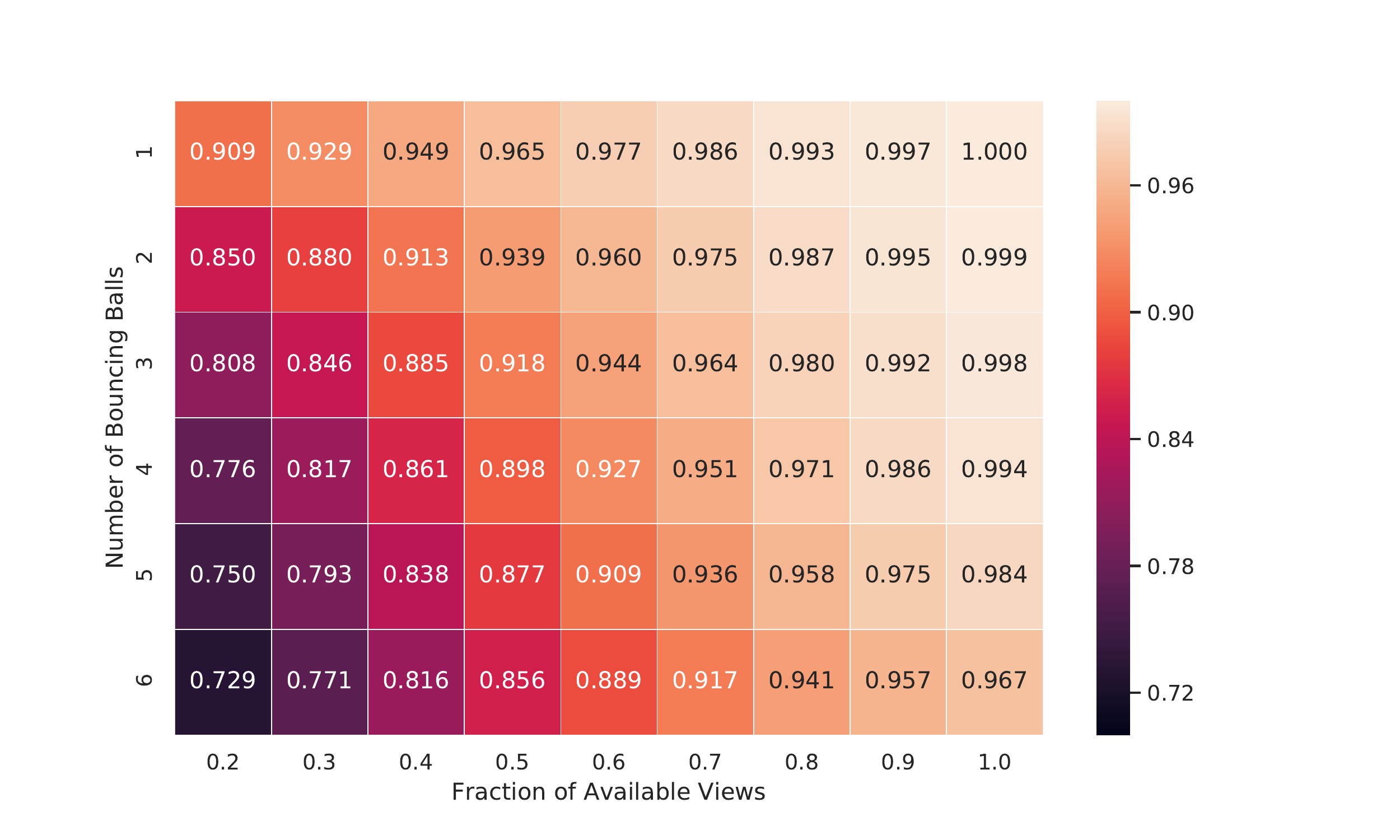}
\caption{TTO}
\label{fig:heatmaps_bacc_tto}
\end{subfigure}

\caption{Balanced accuracy (arithmetic mean of recall and specificity) achieved by all evaluated models for one-step forward prediction task with various number of balls and fractions of available views. All models were trained on video sequences with 3 balls and a constant number of crops / views ($10$ views, corresponding to the right-most columns labelled $1.0$). The color map is consistent across all plots.}
\label{fig:heatmaps_bacc}

\end{figure*}

\begin{figure*}[htp]
\centering

\begin{subfigure}{0.5\textwidth}
\centering
\includegraphics[width=\textwidth]{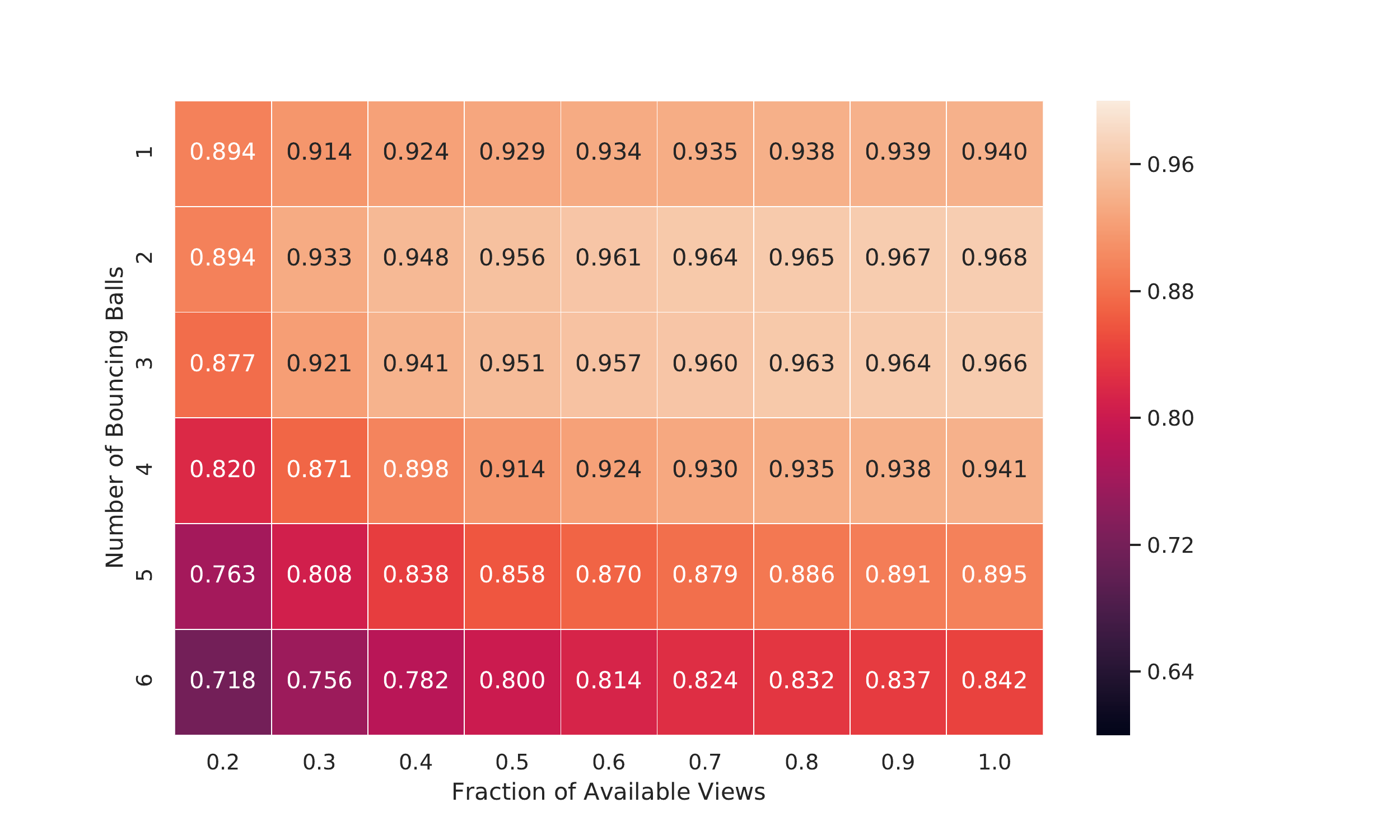}
\caption{S2GRU}
\label{fig:heatmaps_f1_s2gru}
\end{subfigure}\hfill
\begin{subfigure}{0.5\textwidth}
\centering
\includegraphics[width=\textwidth]{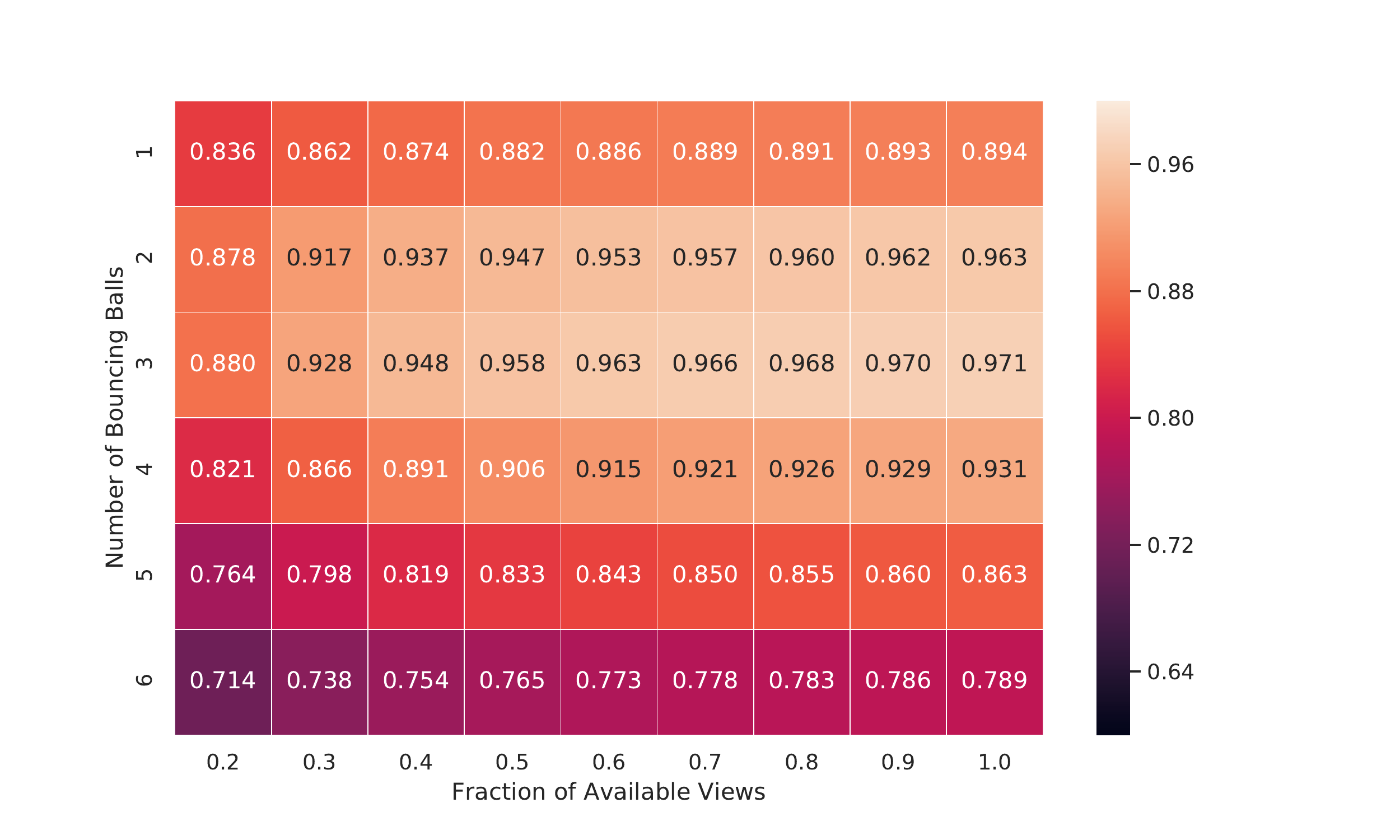}
\caption{RIMs}
\label{fig:heatmaps_f1_rims}
\end{subfigure}

\medskip

\begin{subfigure}{0.5\textwidth}
\centering
\includegraphics[width=\textwidth]{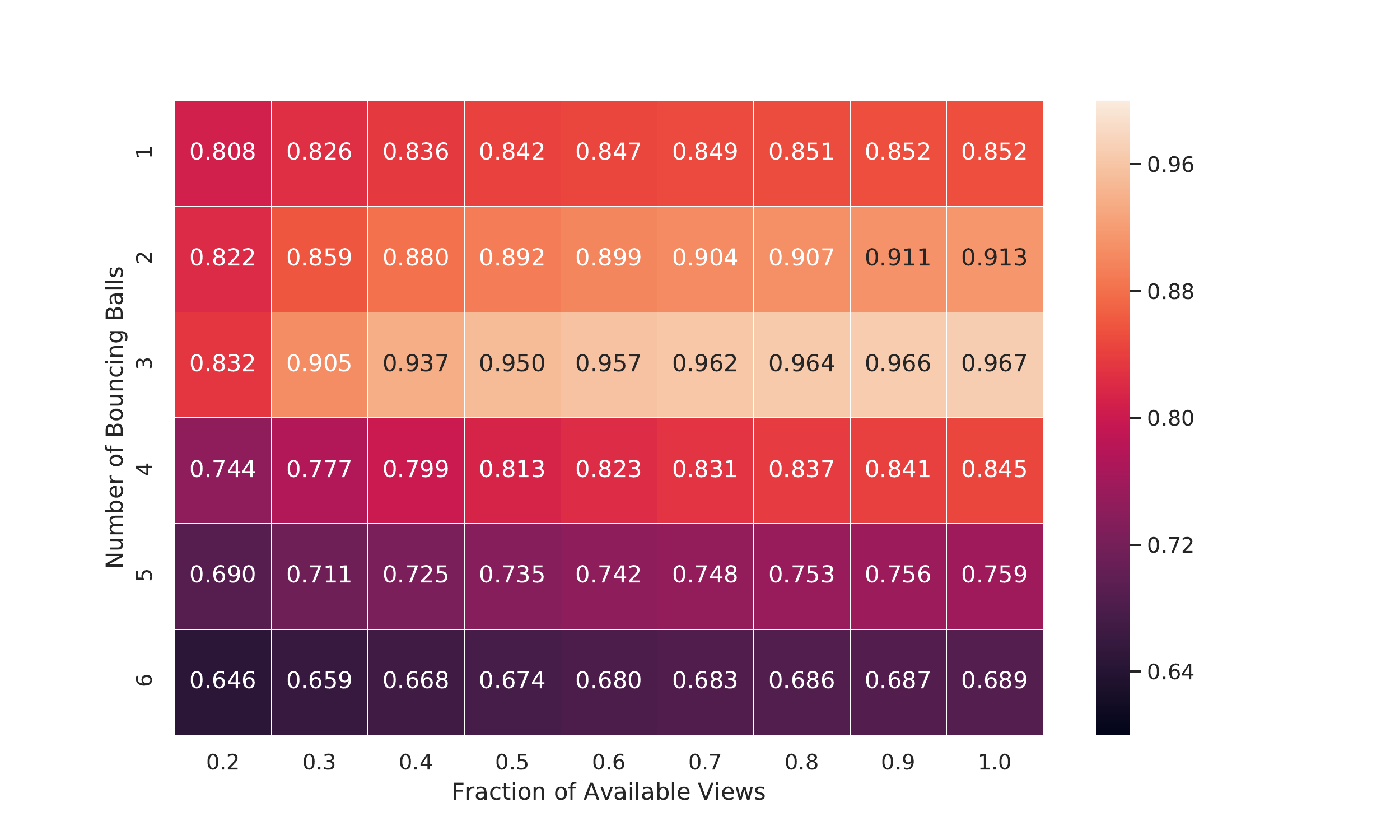}
\caption{RMC}
\label{fig:heatmaps_f1_rmc}
\end{subfigure}\hfill
\begin{subfigure}{0.5\textwidth}
\centering
\includegraphics[width=\textwidth]{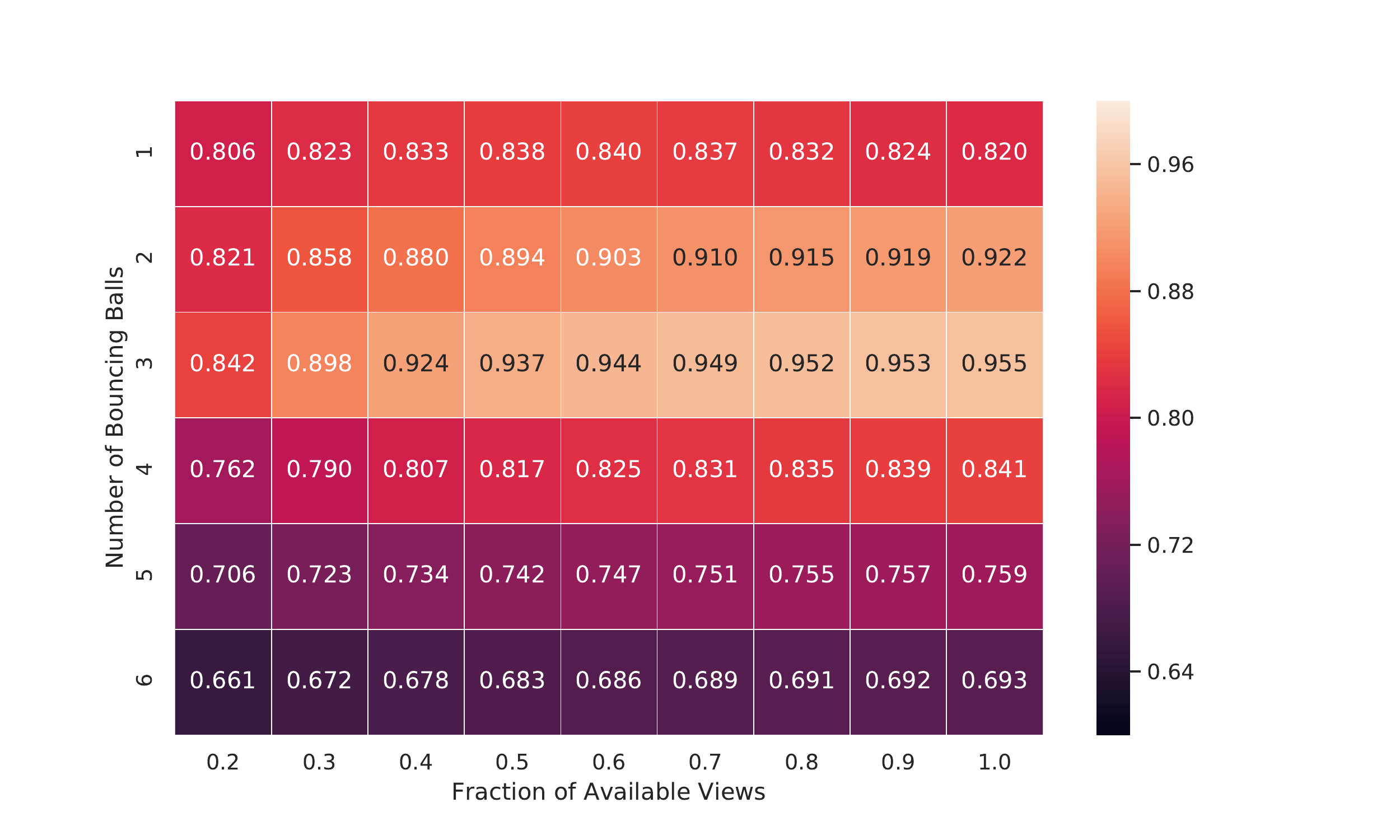}
\caption{LSTM}
\label{fig:heatmaps_f1_lstm}
\end{subfigure}

\medskip

\begin{subfigure}{0.5\textwidth}
\centering
\includegraphics[width=\textwidth]{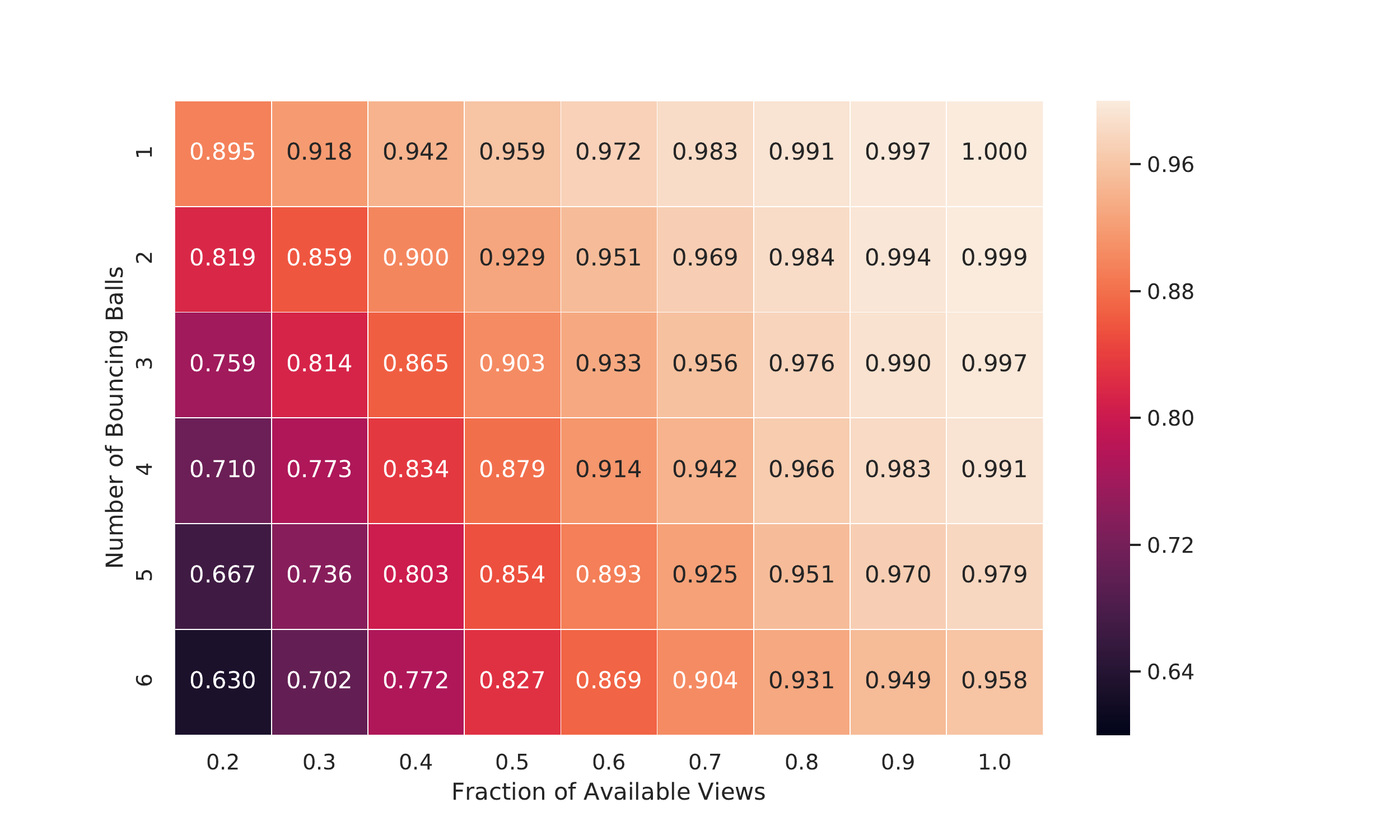}
\caption{TTO}
\label{fig:heatmaps_f1_tto}
\end{subfigure}

\caption{F1-Score (harmonic mean of precision and recall) achieved by all evaluated models for one-step forward prediction task with various number of balls and fractions of available views. All models were trained on video sequences with 3 balls and a constant number of crops / views ($10$ views, corresponding to the right-most columns labelled $1.0$). The color map is consistent across all plots.}
\label{fig:heatmaps_f1}

\end{figure*}

In this section, we evaluate the robustness of all models to dropped crops on in-distribution and OOD data. We measure the performance metrics on one-step forward prediction task on all datasets (with $1$-$6$ balls), albeit by dropping a given fraction of the available input observations.

Figure~\ref{fig:heatmaps_bacc} and \ref{fig:heatmaps_f1} visualize the performance of all evaluated models. We find that S2GRU maintains performance on OOD data even with fewer views (or crops) than it was trained on. Interestingly, we find that the time-travelling oracle (TTO), while robust OOD, is adversely affected by the number of available views. This could be because unlike the other models, it cannot leverage the temporal information to compensate for the missing observations.

\subsection{Starcraft2} \label{app:results_sc2}
\subsubsection{Tabular Results}
The results used to plot Figure~\ref{fig:sc2_robustness} can be found tabulated in Tables~\ref{tab:sc2_fmf1}, \ref{tab:sc2_utf1}, \ref{tab:sc2_hecsnmse} and \ref{tab:sc2_loss}. 
\begin{table*}[]
    \centering
    \begin{tabular}{cccccc}
    \toprule
    \textbf{Model} &  LSTM &       RMC &    S2GRU  &    S2RMC  &       TTO \\
    \textbf{\% of Active Agents} & & & & & \\
    \midrule
    20\%        &           0.570565 &           0.586541 &  \textbf{0.642292} &           0.637618 &           0.550806 \\
    30\%        &           0.599391 &           0.606114 &  \textbf{0.660127} &           0.653950 &           0.578965 \\
    40\%        &           0.630606 &           0.640435 &  \textbf{0.678752} &           0.671476 &           0.605867 \\
    50\%        &           0.638374 &           0.657472 &  \textbf{0.688528} &           0.685988 &           0.627444 \\
    60\%        &           0.681040 &           0.704552 &  \textbf{0.713851} &           0.708786 &           0.671961 \\
    70\%        &           0.709861 &  \textbf{0.737436} &           0.734256 &           0.727980 &           0.723238 \\
    80\%        &           0.721041 &  \textbf{0.748138} &           0.738611 &           0.732114 &           0.740936 \\
    90\%        &           0.750449 &  \textbf{0.778647} &           0.755476 &           0.747613 &           0.786931 \\
    100\%       &           0.765592 &  \textbf{0.795049} &           0.763126 &           0.754637 &           0.813504 \\
    \bottomrule
    \end{tabular}
    \caption{Friendly marker F1 scores on the validation set of the training distribution. Larger numbers are better.}
    \label{tab:sc2_fmf1}
\end{table*}
\begin{table*}[]
    \centering
    \begin{tabular}{cccccc}
    \toprule
    \textbf{Model} &  LSTM &       RMC &    S2GRU  &    S2RMC  &       TTO \\
    \textbf{\% of Active Agents} & & & & & \\
    \midrule
    20\%        &           0.323482 &           0.326685 &  \textbf{0.435318} &           0.377538 &           0.297192 \\
    30\%        &           0.345108 &           0.350621 &  \textbf{0.491934} &           0.433945 &           0.323736 \\
    40\%        &           0.373612 &           0.387733 &  \textbf{0.540163} &           0.485278 &           0.350733 \\
    50\%        &           0.385550 &           0.406048 &  \textbf{0.552589} &           0.510371 &           0.371088 \\
    60\%        &           0.430793 &           0.481986 &  \textbf{0.599470} &           0.566149 &           0.435724 \\
    70\%        &           0.497964 &           0.590214 &  \textbf{0.635928} &           0.606039 &           0.539652 \\
    80\%        &           0.579952 &           0.649277 &  \textbf{0.650682} &           0.623040 &           0.617973 \\
    90\%        &           0.657643 &  \textbf{0.694158} &           0.675294 &           0.655581 &           0.699008 \\
    100\%       &           0.677952 &  \textbf{0.715929} &           0.689669 &           0.672186 &           0.737745 \\
    \bottomrule
    \end{tabular}
    \caption{Unit-type marker (macro averaged) F1 scores on the validation set of the training distribution. Larger numbers are better.}
    \label{tab:sc2_utf1}
\end{table*}
\begin{table*}[]
    \centering
    \begin{tabular}{cccccc}
    \toprule
    \textbf{Model} &  LSTM &       RMC &    S2GRU  &    S2RMC  &       TTO \\
    \textbf{\% of Active Agents} & & & & & \\
    \midrule
    20\%        &          -0.014035 &          -0.013569 & \textbf{-0.011491} &          -0.011921 &          -0.014174 \\
    30\%        &          -0.013355 &          -0.012747 & \textbf{-0.010631} &          -0.011101 &          -0.013539 \\
    40\%        &          -0.012567 &          -0.011808 & \textbf{-0.009906} &          -0.010367 &          -0.012916 \\
    50\%        &          -0.012220 &          -0.011305 & \textbf{-0.009637} &          -0.009887 &          -0.012481 \\
    60\%        &          -0.010888 &          -0.009799 & \textbf{-0.008751} &          -0.009034 &          -0.010929 \\
    70\%        &          -0.009738 &          -0.008469 & \textbf{-0.008068} &          -0.008359 &          -0.009184 \\
    80\%        &          -0.009081 &          -0.008027 & \textbf{-0.007873} &          -0.008162 &          -0.008466 \\
    90\%        &          -0.007970 & \textbf{-0.007180} &          -0.007347 &          -0.007615 &          -0.007038 \\
    100\%       &          -0.007638 & \textbf{-0.006823} &          -0.007103 &          -0.007362 &          -0.006401 \\
    \bottomrule
    \end{tabular}
    \caption{HECS Negative MSE on the validation set of the training distribution. Larger numbers are better.}
    \label{tab:sc2_hecsnmse}
\end{table*}
\begin{table*}[]
    \centering
    \begin{tabular}{cccccc}
    \toprule
    \textbf{Model} &  LSTM &       RMC &    S2GRU  &    S2RMC  &       TTO \\
    \textbf{\% of Active Agents} & & & & & \\
    \midrule
    20\%        &          -0.303051 &          -0.300892 & \textbf{-0.141989} &          -0.146553 &          -0.434099 \\
    30\%        &          -0.258878 &          -0.256878 & \textbf{-0.126037} &          -0.137025 &          -0.347899 \\
    40\%        &          -0.216924 &          -0.211048 & \textbf{-0.113317} &          -0.126882 &          -0.276596 \\
    50\%        &          -0.206582 &          -0.191644 & \textbf{-0.108643} &          -0.113293 &          -0.245019 \\
    60\%        &          -0.158170 &          -0.142643 & \textbf{-0.094380} &          -0.099989 &          -0.175233 \\
    70\%        &          -0.126446 &          -0.109634 & \textbf{-0.084129} &          -0.089527 &          -0.120694 \\
    80\%        &          -0.111735 &          -0.099229 & \textbf{-0.081624} &          -0.086723 &          -0.104135 \\
    90\%        &          -0.082463 &          -0.074518 & \textbf{-0.073439} &          -0.078197 &          -0.071243 \\
    100\%       &          -0.070488 & \textbf{-0.063183} &          -0.069856 &          -0.074041 &          -0.057276 \\
    \bottomrule
    \end{tabular}
    \caption{Log Likelihood (negative loss) on the validation set of the training distribution. Larger numbers are better.}
    \label{tab:sc2_loss}
\end{table*}

\end{appendix}

\end{document}